\documentclass{article}

\usepackage{microtype}
\usepackage{graphicx}
\usepackage{subfigure}
\usepackage{booktabs} % for professional tables

\usepackage{amsfonts}
\usepackage{amsmath} % \begin{align*}
\usepackage[dvipsnames]{xcolor}
\usepackage{soul}
\usepackage{graphicx}
\usepackage{comment}
\usepackage{multirow}

\usepackage{hyperref}

\usepackage[accepted,nohyperref]{icml2018}

\usepackage{xcolor} %hilight

\newcommand{\todo}[2][]{%
    \textcolor{red}{[TODO\ifthenelse{\equal{#1}{}}{}{ (#1)}: #2]}%
}

\newcommand\ie{\emph{i.e.}}

\usepackage{etoolbox}
\makeatletter
\patchcmd\@combinedblfloats{\box\@outputbox}{\unvbox\@outputbox}{}{%
  \errmessage{\noexpand\@combinedblfloats could not be patched}%
}%
\makeatother
\icmltitlerunning{Gated Path Planning Networks}

\begin{document}

\twocolumn[
\icmltitle{Gated Path Planning Networks}

% It is OKAY to include author information, even for blind
% submissions: the style file will automatically remove it for you
% unless you've provided the [accepted] option to the icml2018
% package.

% List of affiliations: The first argument should be a (short)
% identifier you will use later to specify author affiliations
% Academic affiliations should list Department, University, City, Region, Country
% Industry affiliations should list Company, City, Region, Country

% You can specify symbols, otherwise they are numbered in order.
% Ideally, you should not use this facility. Affiliations will be numbered
% in order of appearance and this is the preferred way.
\icmlsetsymbol{equal}{*}

\begin{icmlauthorlist}
\icmlauthor{Lisa Lee}{equal,cmu}
\icmlauthor{Emilio Parisotto}{equal,cmu}
\icmlauthor{Devendra Singh Chaplot}{cmu}
\icmlauthor{Eric Xing}{cmu}
\icmlauthor{Ruslan Salakhutdinov}{cmu}
\end{icmlauthorlist}

\icmlaffiliation{cmu}{Carnegie Mellon University, Machine Learning Department}

\icmlcorrespondingauthor{Lisa Lee}{lslee@cs.cmu.edu}
\icmlcorrespondingauthor{Emilio Parisotto}{eparisot@cs.cmu.edu}

% You may provide any keywords that you
% find helpful for describing your paper; these are used to populate
% the "keywords" metadata in the PDF but will not be shown in the document
\icmlkeywords{Machine Learning, ICML}

\vskip 0.3in
]

% this must go after the closing bracket ] following \twocolumn[ ...

% This command actually creates the footnote in the first column
% listing the affiliations and the copyright notice.
% The command takes one argument, which is text to display at the start of the footnote.
% The \icmlEqualContribution command is standard text for equal contribution.
% Remove it (just {}) if you do not need this facility.

%\printAffiliationsAndNotice{}  % leave blank if no need to mention equal contribution
\printAffiliationsAndNotice{\icmlEqualContribution} % otherwise use the standard text.

%!TEX root = ../main.tex

\begin{abstract}
  Value Iteration Networks (VINs) are effective differentiable path planning modules that can be used by agents to perform navigation while still maintaining end-to-end differentiability of the entire architecture. Despite their effectiveness, they suffer from several disadvantages including training instability, random seed sensitivity, and other optimization problems. In this work, we reframe VINs as recurrent-convolutional networks which demonstrates that VINs couple recurrent convolutions with an unconventional max-pooling activation. From this perspective, we argue that standard gated recurrent update equations could potentially alleviate the optimization issues plaguing VIN. The resulting architecture, which we call the Gated Path Planning Network, is shown to empirically outperform VIN on a variety of metrics such as learning speed, hyperparameter sensitivity, iteration count, and even generalization. Furthermore, we show that this performance gap is consistent across different maze transition types, maze sizes and even show success on a challenging 3D environment, where the planner is only provided with first-person RGB images. 
\end{abstract}

%!TEX root = ../main.tex

\section{Introduction}

A common type of sub-task that arises in various reinforcement learning domains is path finding: finding a shortest set of actions to reach a subgoal from some starting state. Path finding is a fundamental part of any application which requires navigating in an environment, such as robotics~\citep{roomba} and video game AI~\citep{silver2005cooperative}. Due to its ubiquity in these important applications, recent work~\cite{vin} has designed a differentiable sub-module that performs path-finding as directed by the agent in some inner loop. These \emph{Value Iteration Network (VIN)} modules mimic the application of Value Iteration on a 2D grid world, but without a pre-specified model or reward function. VINs were shown to be capable of computing near-optimal paths in 2D mazes and 3D landscapes where the transition model $P(s'|s,a)$ was not provided a priori and had to be learned.

In this paper, we show that VINs are often plagued by training instability, oscillating between high and low performance between epochs; random seed sensitivity, often converging to different performances depending on the random seed that was used; and hyperparameter sensitivity, where relatively small changes in hyperparameters can cause diverging behaviour. Owing to these optimization difficulties, we reframe the VIN as a recurrent-convolutional network, which enables us to replace the unconventional recurrent VIN update (convolution \& max-pooling) with well-established gated recurrent operators such as the LSTM update~\cite{hochreiter1997long}. These \emph{Gated Path Planning Networks (GPPNs)} are a more general model that relaxes the architectural inductive bias of VINs that was designed to perform a computation resembling value-iteration. 

We then establish empirically that GPPNs perform better or equal to the performance of VINs on a wide variety of 2D maze experiments, including different transition models, maze sizes and different training dataset sizes. We further demonstrate that GPNNs exhibit fewer optimization issues than VINs, including reducing random seed and hyperparameter sensitivity and increasing training stability. GPPNs are also shown to work with larger kernel sizes, often outperforming VINs with significantly fewer recurrent iterations, and also learn faster on average and generalize better given less training samples. Finally, we present results for both VIN and GPPN on challenging 3D ViZDoom environments \cite{Kempka2016ViZDoom}, where the planner is only provided with first-person RGB images instead of the top-down 2D maze design.

%!TEX root = ../main.tex

\section{Background}

In reinforcement learning, the environment is formulated as a Markov decision process (MDP) consisting of states $s$, actions $a$, a reward function $\mathcal{R}$, and state transition kernels $P(s' \mid s, a)$. \emph{Value iteration} is a method of computing an optimal policy $\pi$ and its value $V^\pi (s) = \mathbb{E}^\pi \left[ \sum_{t=0}^\infty \gamma^t \mathcal{R}(s_t, a_t,s_{t+1}) \mid s_0 = s \right]$, where $\gamma \in [0, 1]$ is a discount factor and $\mathcal{R}(s_t, a_t,s_{t+1})$ is a reward function. More specifically, value iteration starts with an arbitrary function $V^{(0)}$ and iteratively computes:
\begin{align*}
Q^{(k)}(s, a)
&= \sum_{s'} P(s' \mid s, a) \left( \mathcal{R}(s, a, s') + \gamma V^{(k-1)}(s') \right), \\
V^{(k)}(s)
&= \max_a Q^{(k)}(s, a).
\end{align*}

The value function $V^{(k)}$ converges to $V^*$ in the limit as $k \rightarrow~\infty$, and the optimal policy can be recovered as $\pi^*(s) := \arg\max_a Q^{(\infty)}(s, a)$~\cite{Sutton:1998:IRL:551283}.

Despite the theoretical guarantees, value iteration requires a pre-specified environment model. \citet{vin} introduced the \emph{Value Iteration Network (VIN)}, which is capable of learning these MDP parameters from data automatically. The VIN reformulates value iteration as a recursive process of applying convolutions and max-pooling over the feature channels:
\begin{align}
    \bar{Q}^{(k)}_{\bar{a}, i', j'}
    &= \sum_{i, j} \left(W_{\bar{a}, i,j}^{R} \bar{R}_{i'-i, j'-j} + W_{\bar{a},i,j}^{V} \bar{V}^{(k-1)}_{i'-i,j'-j}\right),
    \nonumber
    \\
    \bar{V}^{(k)}_{i,j}
    &= \max_{\bar{a}} \bar{Q}^{(k)}_{\bar{a}, i, j},
    \label{eq:vin-update}
\end{align}
where the indices $i, j \in [m]$ correspond to cells in the $m \times m$ maze, $\bar{R},\bar{Q},\bar{V}$ is the VIN estimated reward, action-value and value functions, respectively, $\bar{a}$ is the action index of the $\bar{Q}$ feature map, and $W^R, W^V$ are the convolutional weights for the reward function and value function, respectively. In the following iteration, the previous value $\bar{V}$ is stacked with $\bar{R}$ for the convolution step.

\citet{vin} showed that VINs have much greater success at path planning than baseline CNN and feedforward architectures in a variety of 2D and graph-based navigation tasks. The demonstrated success of VIN has made it an important component of models designed to solve downstream tasks where navigation is crucial~\cite{qmdpnet,cogmap,unifyingmap}. For example, \citet{cogmap,unifyingmap} designed a Deep RL agent to perform navigation within partially observable and noisy environments by combining a VIN module with a 2D-structured memory map.

%!TEX root = ../main.tex

\section{Method}
\label{section:method}

In this work, we explore whether the inductive biases provided by the VIN are even necessary: is it possible that using alternative, more general architectures might work better than those of the VIN? We can view the VIN update~\eqref{eq:vin-update} within the perspective of a convolutional-recurrent network, updating a recurrent state $V^{(k)}_{i',j'}$ at every spatial position $(i',j')$ in each iteration:
\begin{align}
  \bar{V}^{(k)}_{i',j'}
  &= \max_{\bar{a}} \left(\sum_{i, j} W_{\bar{a}, i,j}^{R} \bar{R}_{i'-i, j'-j} + W_{\bar{a},i,j}^{V} \bar{V}^{(k-1)}_{i'-i,j'-j} \right) \nonumber\\
    &= \max_{\bar{a}} \left(W_{\bar{a}}^{R} \bar{R}_{[i',j',3]} + W_{\bar{a}}^V \bar{V}^{(k-1)}_{[i',j',3]} \right),
    \label{eq:vin-crn-update}
\end{align}
where $X_{[i',j',F]}$ denotes the image patch centered at position $(i',j')$ with kernel size $F$. From~\eqref{eq:vin-crn-update}, it can be seen that VIN follows the standard recurrent neural network (RNN) update where the recurrent state is updated by taking a linear combination of the input $\bar{R}$ and the previous recurrent state  $\bar{V}^{(k-1)}$, and passing their sum through a nonlinearity $\max_{\bar{a}}$. The main differences from a standard RNN are the following: the non-conventional nonlinearity (channel-wise max-pooling) used in VIN; the hidden dimension of the recurrent network, which is essentially one; the sparse weight matrices, where the non-zero values of the weight matrices represent neighboring inputs and units which are local in space; and the restriction of kernel sizes to 3.

Under this perspective, it is easy to question whether the adherence to these strict architectural biases is even necessary, given the long history of demonstrations that standard non-gated recurrent operators are difficult to optimize due to effects such as vanishing and exploding gradients~\cite{PascanuMB13}.

We can easily replace the recurrent VIN update in~\eqref{eq:vin-crn-update} with the well-established LSTM update~\cite{hochreiter1997long}, whose gated update alleviates many of the problems with standard recurrent networks:

\begin{footnotesize}
\begin{align}
  &h^{(k)}_{i',j'}, c^{(k)}_{i',j'} = \nonumber \\
   &\quad\textbf{LSTM} \left( \sum_{\bar{a}} \left( W_{\bar{a}}^R \bar{R}_{[i',j',F]} + W_{\bar{a}}^h h^{(k-1)}_{[i',j',F]} \right), c^{(k-1)}_{i',j'}\right),
  \label{eq:GPPN-update}
\end{align}
\end{footnotesize}

where $F$ is the convolution kernel size. This recurrent update~\eqref{eq:GPPN-update} still maintains the convolutional properties of the input and recurrent weight matrix as in VIN. It involves taking as input the $F \times F$ convolution of the input vector $\bar{R}$ and previous hidden states $h^{(k-1)}$, and the previous cell state $c_{i',j'}^{(k-1)}$ of the LSTM at the central position $(i', j')$. We call path planning modules which use these gated updates \emph{Gated Path Planning Networks (GPPNs)}. The GPPN is an LSTM which uses convolution of previous spatially-contiguous hidden states for its input.

\section{Environments and Maze Transition Types}

We test VIN and GPPN on 2D maze environments and 3D ViZDoom environments (Figure~\ref{fig:doom}) on a variety of settings such as training dataset size, maze size and maze transition kernel.

We used three different maze transition kernels: In \textbf{NEWS}, the agent can move North, East, West, or South; in \textbf{Differential Drive}, the agent can move forward along its current orientation, or turn left/right by 90 degrees; in \textbf{Moore}, the agent can move to any of the eight cells in its Moore neighborhood. In the NEWS and Moore transition types, the target is an x-y coordinate, while in Differential Drive the target contains an orientation along with the x-y coordinate. Consequently, the dimension of the goal map given as input to the models is $1 \times m \times m$ for NEWS and Moore, and $4 \times m \times m$ for Differential Drive, where $m$ is the maze size. 

\subsection{2D Maze Environment}
\begin{figure}[t]
\subfigure[2D maze]{\includegraphics[height=80pt, trim={104pt 37pt 95pt 39pt}, clip]{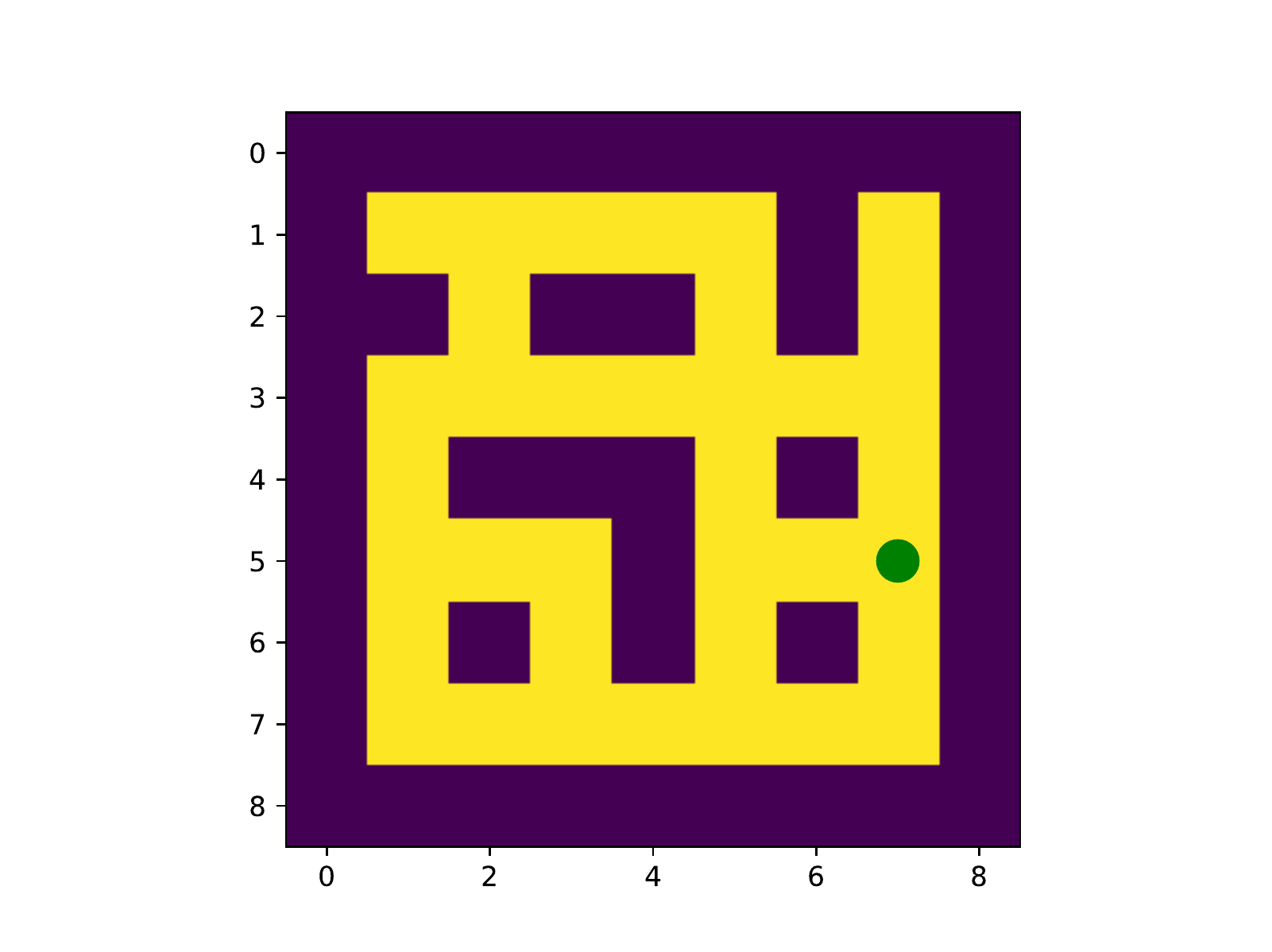}}
\label{fig:2d-maze-example}
\hspace{20pt}
\subfigure[3D ViZDoom maze] {\includegraphics[height=80pt]{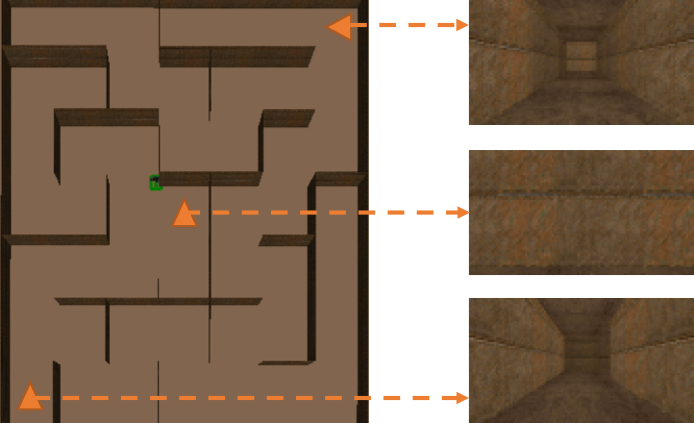}}
\label{fig:3d-maze-example}
\caption{(a) A sample 2D maze. (b) A sample 3D Doom maze and examples of screenshots showing the first-person view of the environment at three locations.} \label{fig:doom}

\end{figure}

The 2D maze environment is created with a maze generation process that uses Depth-First Search with the Recursive Backtracker algorithm~\cite{wiki} to construct the maze tree, resulting in a fully connected maze (see Figure~\ref{fig:doom}a). For each maze, we sample a probability $d$ uniformly from [0,1]. Then for each wall, we delete the wall with probability $d$. 

For our experiments on the 2D mazes, the state vector consists of the maze and the goal location, each of which are represented by a binary $m \times m$ matrix, where $m \times m$ is the maze size. We use early stopping based on validation set metrics to choose the final models.

%!TEX root = ../main.tex

\subsection{3D ViZDoom Environment}

We use the Doom Game Engine and the ViZDoom API \cite{Kempka2016ViZDoom} to create mazes in a simulated 3D environment (see Figure~\ref{fig:doom}b). The maze design for the 3D mazes are generated in exactly the same manner as the 2D mazes, using Depth-First Search with the Recursive Backtracker algorithm followed by wall pruning with a uniformly sampled probability $d$.  For each Doom maze, we take RGB screenshots showing the first-person view of the environment at each position and orientation. A sample 3D Doom maze and example screenshot images are shown in Figure~\ref{fig:doom}. For an $m \times m$ maze with 4 orientations, this results in a total of $4 m^2$ images. 

In the 3D ViZDoom experiments, these map images are given as input to the model (instead of the 2D map design). This setup is similar to the one used for localization experiments by \citet{chaplot2018active} who argue that these images are easier to obtain as compared to constructing an accurate map design of an environment in the real world. The model needs to learn to infer the map design from these images along with learning to plan, which makes the task more challenging in 3D environments.

%!TEX root = ../main.tex

\section{Experiments \& Discussion}\label{section:experiments}

In this section, we empirically compare VIN and GPPN using two metrics: \textbf{\%Optimal} (\%Opt) is the percentage of states whose predicted paths under the policy estimated by the model has optimal length, and \textbf{\%Success} (\%Suc) is the percentage of states whose predicted paths under the policy estimated by the model reach the goal state. The reported performance is on a held-out test split. In contrast with the metrics reported in~\cite{vin}, we do not stochastically sample rollouts but instead evaluate and train the output policy of the models directly on all states simultaneously. This reduces optimization noise and makes it easier to tell whether difficulties with training are due to sampling noise or model architecture/capacity. 

All analyses are based on 2D maze results, except in Section~\ref{section:3d-ViZDoom} where we discuss 3D ViZDoom results. In order to make comparison fair, we utilized a hidden dimension of 150 for GPPN and 600 for VIN, owing to the approximately 4$\times$ increase in parameters a GPPN contains due to the 4 gates it computes. Unless otherwise noted, the results were obtained by doing a hyperparameter sweep of $(K, F)$ over $K \in \{5, 10, 15, 20, 30\}$ and $F \in \{3, 5, 7, 9, 11\}$, and using a 25k/5k/5k train-val-test split. Other experimental details are deferred to the Appendix.

%!TEX root = ../main.tex
\begin{table}[t]
  \centering\footnotesize
\caption{Test performance on 2D mazes of size $15 \times 15$ with \textbf{varying kernel sizes} $F$ and best $K$ setting for each $F$. Bold indicates best result across all $F$ for each model and transition kernel. VIN performs worse with larger $F$, while GPPN is more robust when $F$ is varied and actually works better with larger $F$. }
  \setlength\tabcolsep{3pt}
  \begin{tabular}{cc|cc|cc|cc}
    \toprule
    & & \multicolumn{2}{c|}{NEWS} & \multicolumn{2}{c|}{Moore} & \multicolumn{2}{c}{Diff. Drive}\\
    Model & F & \%Opt & \%Suc & \%Opt & \%Suc & \%Opt & \%Suc \\
    \toprule
VIN & 3 & 93.4 & 93.5 & 90.5 & 91.3 & \textbf{98.4} & \textbf{99.1}\\
VIN & 5 & \textbf{93.9} & \textbf{94.1} & \textbf{96.3} & \textbf{96.6} & 96.4 & 98.6\\
VIN & 7 & 92.7 & 93.0 & 95.1 & 95.6 & 92.2 & 96.2\\
VIN & 9 & 86.8 & 87.8 & 92.0 & 93.0 & 91.2 & 95.2\\
VIN & 11 & 87.6 & 88.3 & 92.7 & 93.8 & 87.9 & 93.8\\
\midrule
GPPN & 3 & 97.6 & 98.3 & 96.8 & 97.6 & 96.4 & 98.1\\
GPPN & 5 & 98.6 & 99.0 & 98.4 & 99.1 & 98.7 & 99.5\\
GPPN & 7 & 99.0 & 99.3 & 98.8 & 99.3 & 99.1 & 99.7\\
GPPN & 9 & 99.0 & 99.4 & \textbf{98.8} & \textbf{99.3} & \textbf{99.3} & \textbf{99.7}\\
GPPN & 11 & \textbf{99.2} & \textbf{99.5} & 98.6 & 99.2 & 99.2 & 99.6\\
    \bottomrule
    \end{tabular}
\label{table:2d-varying-f}
\end{table}

%!TEX root = ../main.tex
\begin{table*}[t]
  \centering
  \footnotesize
  \caption{Test performance on 2D mazes of size $15 \times 15$ with \textbf{varying kernel sizes $F$} and \textbf{iteration counts $K$}. ``--'' indicates the training diverged. GPPN outperforms VIN under best settings of $(K, F)$, indicated in bold. By utilizing a larger $F$, GPPN can learn to more effectively propagate information spatially in a smaller number of iterations ($K \leq 10$) than VIN can.}
  \vspace{2pt}
  \setlength\tabcolsep{2pt}
  \begin{tabular}{cc|ccccc|ccccc|ccccc}
    \toprule
    &
    & \multicolumn{5}{c|}{\%Opt for NEWS}
    & \multicolumn{5}{c|}{\%Opt for Moore}
    & \multicolumn{5}{c}{\%Opt for Differential Drive} \\
Model & $K$
  & $F=3$ & $F=5$ & $F=7$ & $F=9$ & $F=11$
  & $F=3$ & $F=5$ & $F=7$ & $F=9$ & $F=11$
  & $F=3$ & $F=5$ & $F=7$ & $F=9$ & $F=11$
  \\
    \midrule
VIN & 5 & 55.6 & 87.7 & 84.6 & 86.3 & 86.6 & 75.0 & 86.7 & 88.9 & 92.0 & 92.3 & 74.8 & 91.9 & 91.5 & 91.2 & 87.9\\
VIN & 10 & 79.0 & 83.3 & 92.2 & 86.8 & 86.7 & 90.5 & 91.4 & 95.1 & 89.4 & 92.7 & 92.4 & 96.1 & 92.2 & 84.0 & 64.4\\
VIN & 15 & 91.3 & 92.9 & 92.7 & 85.4 & 87.6 & 88.7 & 89.6 & 92.4 & 90.0 & 91.0 & 96.7 & 96.4 & 90.1 & 65.2 & 23.0\\
VIN & 20 & 93.4 & \textbf{93.9} & 91.4 & 86.3 & 85.5 & 80.9 & 92.8 & 90.7 & 89.1 & 90.4 & 97.7 & 94.8 & 89.0 & 40.0 & 22.3\\
VIN & 30 & 71.2 & 92.8 & 84.5 & 86.5 & 86.4 & 80.5 & \textbf{96.3} & 92.5 & 91.7 & 89.1 & \textbf{98.4} & 95.9 & 89.5 & -- & --\\
\midrule
GPPN & 5 & 66.2 & 86.5 & 90.8 & 92.4 & 93.0 & 75.9 & 90.4 & 93.4 & 93.9 & 94.1 & 62.4 & 82.3 & 88.6 & 90.1 & 91.2\\
GPPN & 10 & 91.2 & 96.1 & 97.1 & 97.6 & 97.7 & 93.3 & 96.5 & 97.4 & 97.6 & 97.4 & 87.7 & 95.4 & 96.1 & 97.0 & 97.4\\
GPPN & 15 & 95.3 & 98.1 & 98.5 & 98.3 & 98.8 & 96.1 & 97.7 & 98.1 & 98.1 & 98.3 & 93.5 & 97.1 & 97.8 & 97.7 & 99.0\\
GPPN & 20 & 97.4 & 98.4 & 99.0 & 99.0 & \textbf{99.2} & 96.8 & 98.4 & 98.5 & 98.7 & 98.6 & 95.8 & 97.9 & 98.4 & 98.4 & 98.9\\
GPPN & 30 & 97.6 & 98.6 & 99.0 & 98.6 & 98.8 & 98.0 & 98.4 & 98.8 & \textbf{98.8} & 98.4 & 96.4 & 98.7 & 99.1 & \textbf{99.3} & 99.2\\
    \bottomrule
    \end{tabular}
\label{table:2d-varying-k-f}
\end{table*}

\subsection{Varying Kernel Size $F$}

One question that can be asked of the architectural choices of the VIN is whether the kernel size needs to be the same dimension as the true underlying transition model. The kernel size used in VIN was set to $3\times 3$ with a stride of $1$, which is sufficient to represent the true transition model when the agent can move anywhere in the Moore neighborhood, but it limits the rate at which information propagates spatially with each iteration. With a kernel size of $3\times 3$ and stride of $1$, the receptive field of a unit in the last iteration's feature map increases with rate $(3+2K)\times (3+2K)$ where $K$ is the iteration count, meaning that the maximum path length information travels scales directly with iteration count $k$. Therefore for long-term planning in larger environments, \citet{vin} designed a multi-scale variant called the Hierarchical VIN. Hierarchical VINs rely on downsampling the maps into multi-scale hierarchies, and then doing VIN planning and up-scaling, progressively growing the map until it regains its original, un-downsampled size. 

Another potential method to do long-range planning without requiring a multi-scale hierarchy is to instead increase the kernel size. An increased kernel size would cause the receptive field to grow more rapidly, potentially allowing the models to require fewer iterations $K$ before reaching well-performing policies. In this section, we sought to test out the feasibility of increasing the kernel size of VINs and GPPNs. These results are summarized in Table~\ref{table:2d-varying-f}. All the models were trained with the best $K$ setting for each $F$ and transition kernel.
From the results, we can clearly see that GPPN can handle training with larger $F$ values, and moreover, GPPN often performs better than VIN with larger values of $F$. In contrast, we can observe that VIN's performance drops significantly after its kernel size is increased more than 5, with its best performing settings being either $3$ or $5$ depending on the true transition model. These results show that GPPN can learn planning approximations that work with $F>3$ much more stably than VIN, and could further suggest that GPPN can work as well as VIN with less iterations.

\subsection{Varying Iteration Count $K$}

Following the above results showing that GPPN benefits from increased $F$, we further evaluated the effect of varying both iteration count $K$ and kernel size $F$ on the VIN and GPPN models. Table \ref{table:2d-varying-k-f} shows \%Optimal and \%Success results of VIN and GPPN on 15$\times$15 2D mazes for different values of $F$ and $K$. We can see from NEWS column in the table that GPPN with $F > 7$ can get results on par with the best VIN model with only $K=5$ iterations. This shows that GPPN can learn to more effectively propagate information spatially in a smaller number of iterations than VIN can, and outperforms VIN even when VIN is given a much larger number of iterations. Additionally, we can see that VIN has significant trouble learning when both $K$ and $F$ are large in the differential drive mazes and to a lesser extent in the NEWS mazes.

Table~\ref{table:2d-varying-k} shows the results of VIN and GPPN with varying iteration counts $K$ and the best $F$ setting for each $K$. Owing to the larger kernel size, GPPN with smaller number of  iterations $K \leq 10$ can get results on par with the best VIN model. Generally, both models benefit from a larger $K$ (assuming the best $F$ setting is used).

%!TEX root = ../main.tex
\begin{table}[t]
  \centering\footnotesize
\caption{Test performance on 2D mazes of size $15 \times 15$ with \textbf{varying iteration counts} $K$ and best $F$ setting for each $K$. Bold indicates best result across all $K$ for each model and transition kernel. Generally, increasing $K$ improves performance. }
\vspace{2pt}
  \setlength\tabcolsep{3pt}
  \begin{tabular}{cc|cc|cc|cc}
    \toprule
    & & \multicolumn{2}{c|}{NEWS} & \multicolumn{2}{c|}{Moore} & \multicolumn{2}{c}{Diff. Drive}\\
    Model & K & \%Opt & \%Suc & \%Opt & \%Suc & \%Opt & \%Suc \\
    \toprule
VIN & 5 & 87.7 & 88.4 & 92.3 & 93.3 & 91.9 & 95.8\\
VIN & 10 & 92.2 & 92.5 & 95.1 & 95.6 & 96.1 & 97.9\\
VIN & 15 & 92.9 & 93.0 & 92.4 & 93.9 & 96.7 & 98.3\\
VIN & 20 & \textbf{93.9} & \textbf{94.1} & 92.8 & 94.0 & 97.7 & 98.8\\
VIN & 30 & 92.8 & 93.2 & \textbf{96.3} & \textbf{96.6} & \textbf{98.4} & \textbf{99.1}\\
\midrule
GPPN & 5 & 93.0 & 94.3 & 94.1 & 96.1 & 91.2 & 95.6\\
GPPN & 10 & 97.7 & 98.4 & 97.6 & 98.4 & 97.4 & 98.8\\
GPPN & 15 & 98.8 & 99.2 & 98.3 & 98.9 & 99.0 & 99.6\\
GPPN & 20 & \textbf{99.2} & \textbf{99.5} & 98.7 & 99.2 & 98.9 & 99.5\\
GPPN & 30 & 99.0 & 99.3 & \textbf{98.8} & \textbf{99.3} & \textbf{99.3} & \textbf{99.7}\\
    \bottomrule
    \end{tabular}
\label{table:2d-varying-k}
\end{table}

\subsection{Different Maze Transition Kernels}

From Tables~\ref{table:2d-varying-f} and ~\ref{table:2d-varying-k}, we can observe the performance of VIN and GPPN across a variety of different underlying groundtruth transition kernels (NEWS, Moore, and Differential Drive). From these results, we can see that GPPN consistently outperforms VIN on all the transition kernel types. An interesting observation is that VIN does very well at Differential Drive, consistently obtaining high results, although GPPN still does better than or on par with VIN. The reasons why VIN is so well suited to Differential Drive are not clear, and a preliminary analysis of VIN's feature weights and reward vectors did not reveal any intuition on why this is the case.

%!TEX root = ../main.tex
\begin{table}[t]
  \centering\footnotesize
\caption{Test performance on 2D mazes of size $15 \times 15$ with \textbf{varying dataset sizes $N$} under best settings of $(K, F)$ for each model. Both models improve with more training data (larger $N$). GPPN performs relatively better than VIN with less data, suggesting that the VIN architectural biases do not help generalization performance.}
\vspace{2pt}
  \setlength\tabcolsep{3pt}
  \begin{tabular}{cc|cc|cc|cc}
    \toprule
    & & \multicolumn{2}{c|}{NEWS} & \multicolumn{2}{c|}{Moore} & \multicolumn{2}{c}{Diff. Drive}\\
     $N$ & Model & \%Opt & \%Suc & \%Opt & \%Suc & \%Opt & \%Suc \\
    \midrule
10k & VIN & 90.3 & 90.6 & 88.1 & 90.5 & 97.5 & 98.4\\
10k & GPPN & \textbf{97.8} & \textbf{98.6} & \textbf{97.6} & \textbf{98.4} & \textbf{98.0} & \textbf{99.4}\\
\midrule
25k & VIN & 93.9 & 94.1 & 96.3 & 96.6 & 98.4 & 99.1\\
25k & GPPN & \textbf{99.2} & \textbf{99.5} & \textbf{98.8} & \textbf{99.3} & \textbf{99.3} & \textbf{99.7}\\
\midrule
100k & VIN & 97.3 & 97.3 & 97.1 & 97.5 & 98.9 & 99.4\\
100k & GPPN & \textbf{99.9} & \textbf{99.9} & \textbf{99.7} & \textbf{99.8} & \textbf{99.9} & \textbf{99.9}\\
    \bottomrule
    \end{tabular}
\label{table:2d-varying-train-size}
\end{table}

\subsection{Effect of Dataset Size}

A potential benefit of the stronger architectural biases of VIN might be that they can enable better generalization given less training data. In this section, we designed experiments that set out to test this hypothesis. We trained VINs and GPPNs on datasets with varying number of training samples for all three maze transition kernels, and the results are given in Table~\ref{table:2d-varying-train-size}. We can see that GPPN consistently outperforms VIN across all dataset sizes and maze models. Interestingly, we can observe that the performance gap between VIN and GPPN is larger the less data there is, demonstrating the opposite effect to our hypothesis. This could suggest that the architectural biases do not in fact aid generalization performance, or that there is another problem, such as perhaps the difficulty of optimizing VIN, that overshadows the benefit that the inductive bias could potentially provide. 

%!TEX root = ../main.tex
\begin{table}[t]
\centering\fontsize{8.5}{10.2}\selectfont
\caption{Mean and standard deviation \%Opt after 30 epochs, taken over 3 runs, on 2D mazes of size $15 \times 15$. Bold indicates best result across all $K$ for each model and transition kernel. The results were obtained using the best setting of $F$ for each $K$ and dataset size 100k. GPPN exhibits lower variance between runs.}
  \setlength\tabcolsep{3pt}
  \vspace{2pt}
  \begin{tabular}{cc|cccc|cccc}
    \toprule
    & & \multicolumn{4}{c|}{NEWS \%Opt} & \multicolumn{4}{c}{Diff. Drive \%Opt}\\
    & & \multicolumn{2}{c}{Train} & \multicolumn{2}{c|}{Val.} & \multicolumn{2}{c}{Train} & \multicolumn{2}{c}{Val.}\\
    Model & $K$ & mean & std & mean & std & mean & std & mean & std \\
    \midrule
VIN & 5 & 90.1 & 0.1 & 90.1 & 1.5 & 88.4 & 1.0 & 95.4 & 1.1\\
VIN & 10 & 92.8 & 0.6 & 92.7 & 1.4 & 92.3 & 0.5 & 93.9 & 0.2\\
VIN & 15 & 93.4 & 1.2 & 94.2 & 0.6 & \textbf{95.8} & \textbf{0.5} & \textbf{97.0} & \textbf{0.3}\\
VIN & 20 & \textbf{93.1} & \textbf{1.5} & \textbf{94.3} & \textbf{0.8} & 96.4 & 0.2 & 96.8 & 1.1\\
\midrule
GPPN & 5 & 95.5 & 0.2 & 95.2 & $<$0.1 & 93.8 & 0.2 & 93.4 & $<$0.1\\
GPPN & 10 & 99.1 & 0.1 & 99.0 & $<$0.1 & 98.7 & 0.1 & 98.2 & 0.2\\
GPPN & 15 & 99.6 & $<$0.1 & 99.6 & $<$0.1 & 99.4 & 0.1 & 99.3 & 0.1\\
GPPN & 20 & \textbf{99.7} & \textbf{$<$0.1} & \textbf{99.7} & \textbf{0.1} & \textbf{99.8} & \textbf{$<$0.1} & \textbf{99.7} & \textbf{0.1}\\
    \bottomrule
    \end{tabular}
\label{table:variance}
\end{table}

%!TEX root = ../main.tex

\begin{figure}[t]
\centering
\includegraphics[width=0.47\textwidth]{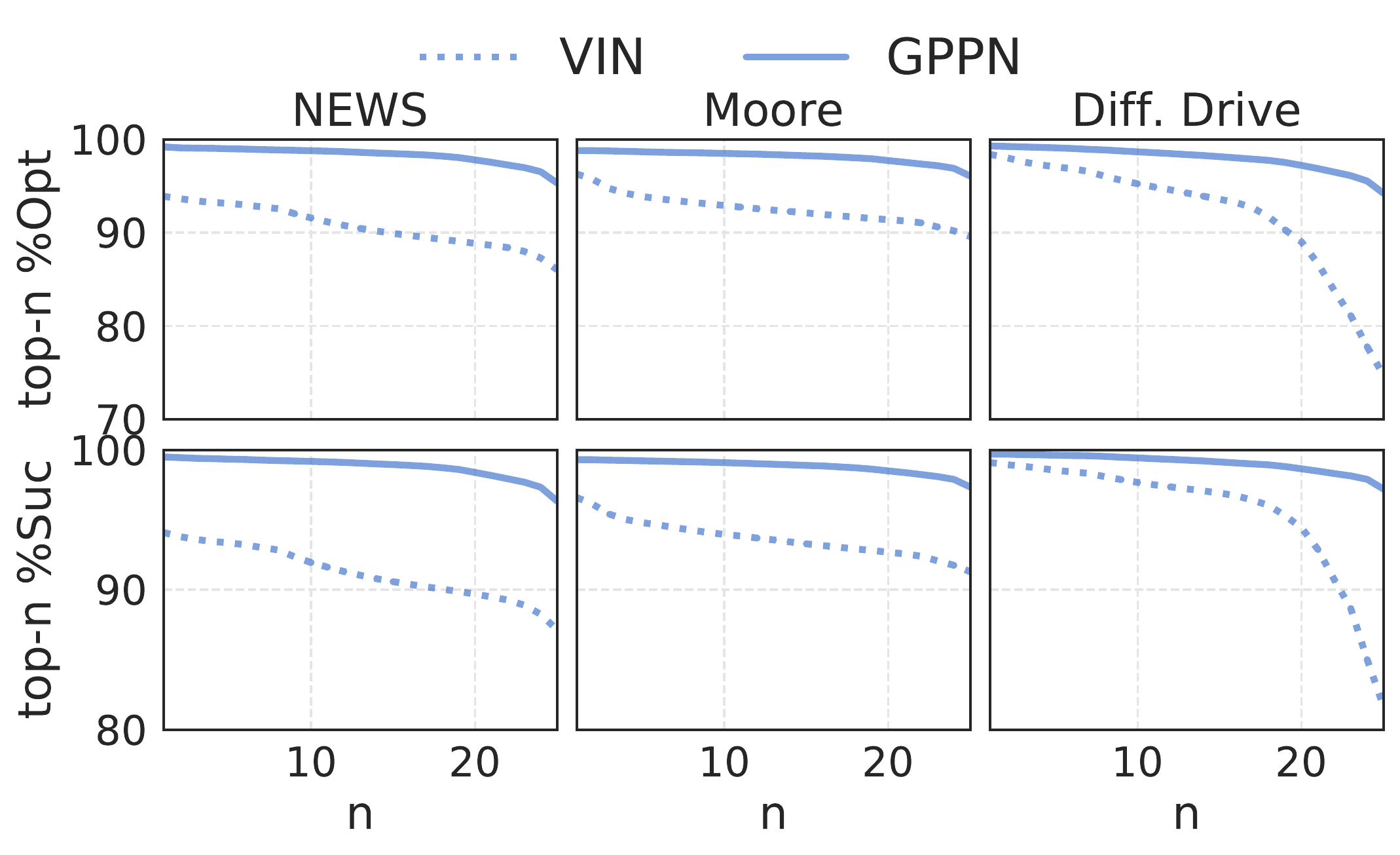}
\vspace{-10pt}
\caption{The y-axis is the average Test \%Opt (or \%Suc) of the top-$n$ hyperparameter settings $(K, F)$ over $K \in \{5, 10, 15, 20, 30\}$ and $F \in \{3, 5, 7, 9, 11\}$. The results are on 2D mazes of size $15 \times 15$. These plots measure how stable the performance of each model is to hyperparameter changes as we increase the number of hyperparameter settings considered. GPPN exhibits less hyperparameter sentivitiy.}
\label{fig:hyperparameter-sensitivity}
\end{figure}

%!TEX root = ../main.tex
\begin{figure*}[t]
\newcommand{\figheight}{125pt}
\centering
\includegraphics[width=\textwidth]{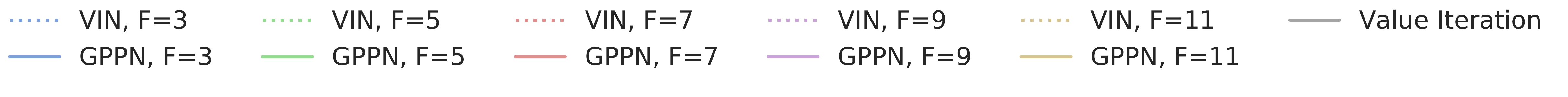}

\includegraphics[height=\figheight]{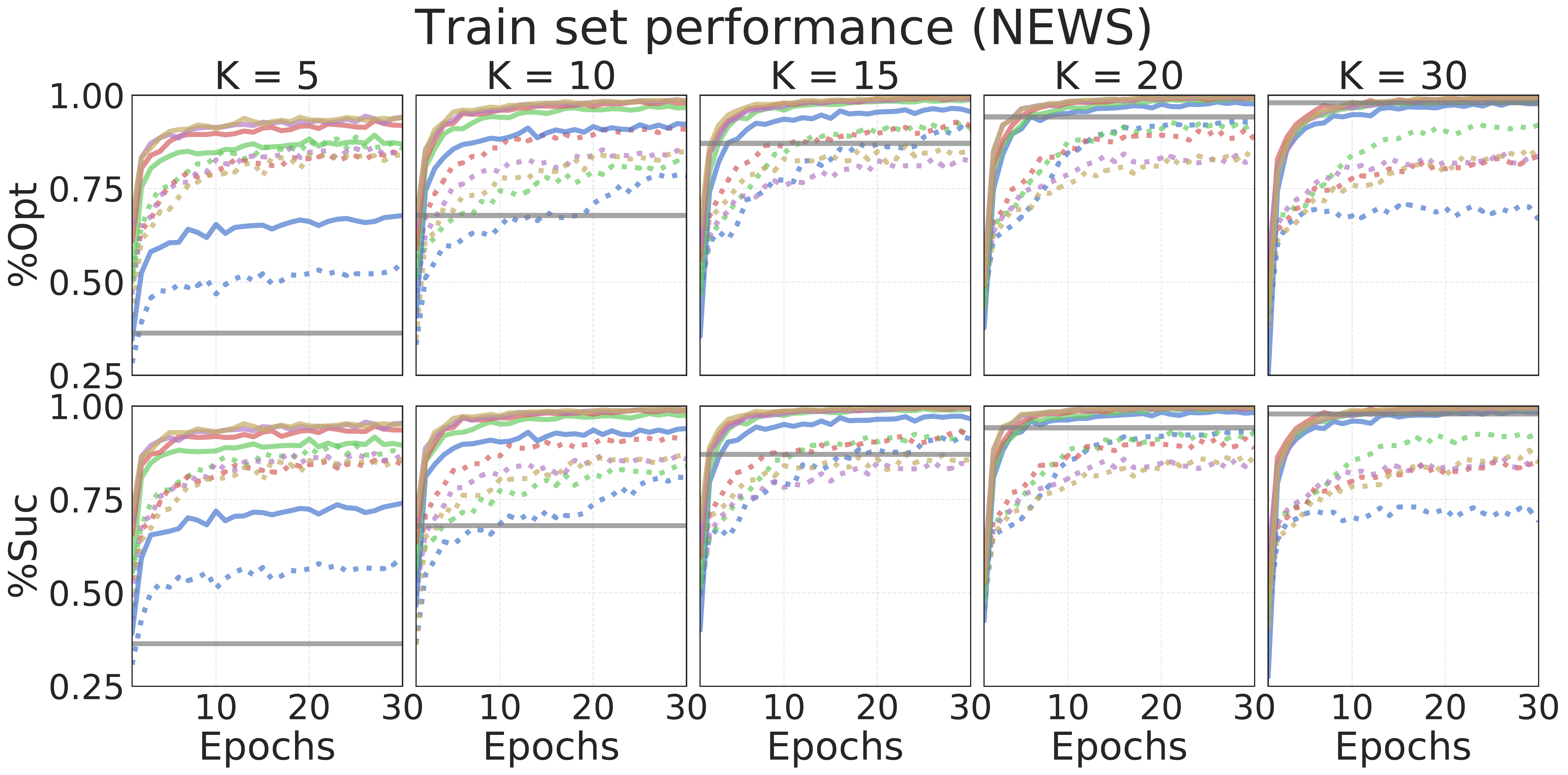}
\includegraphics[height=\figheight, trim={125pt 0 0 0}, clip]{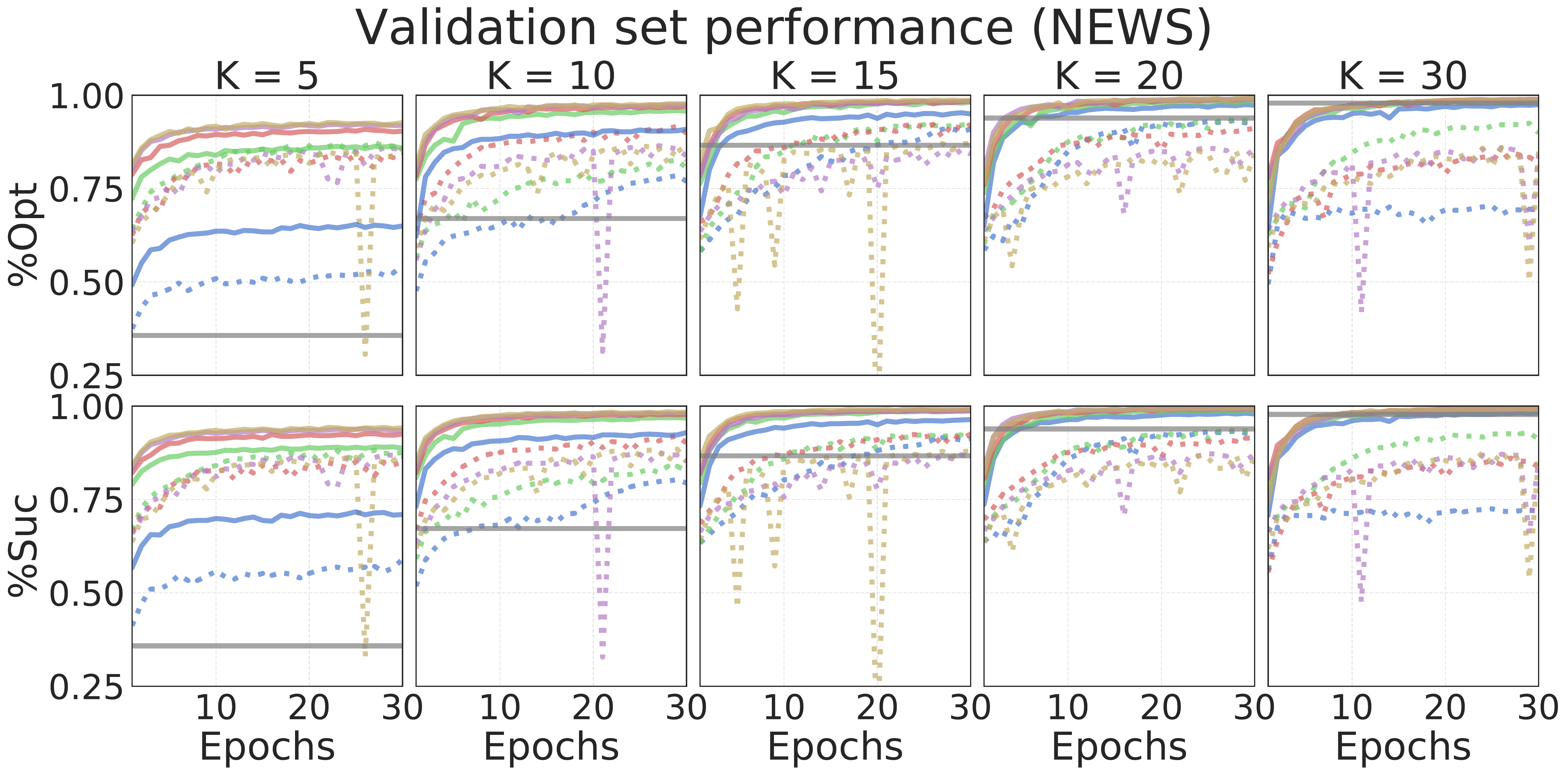}

\includegraphics[height=\figheight]{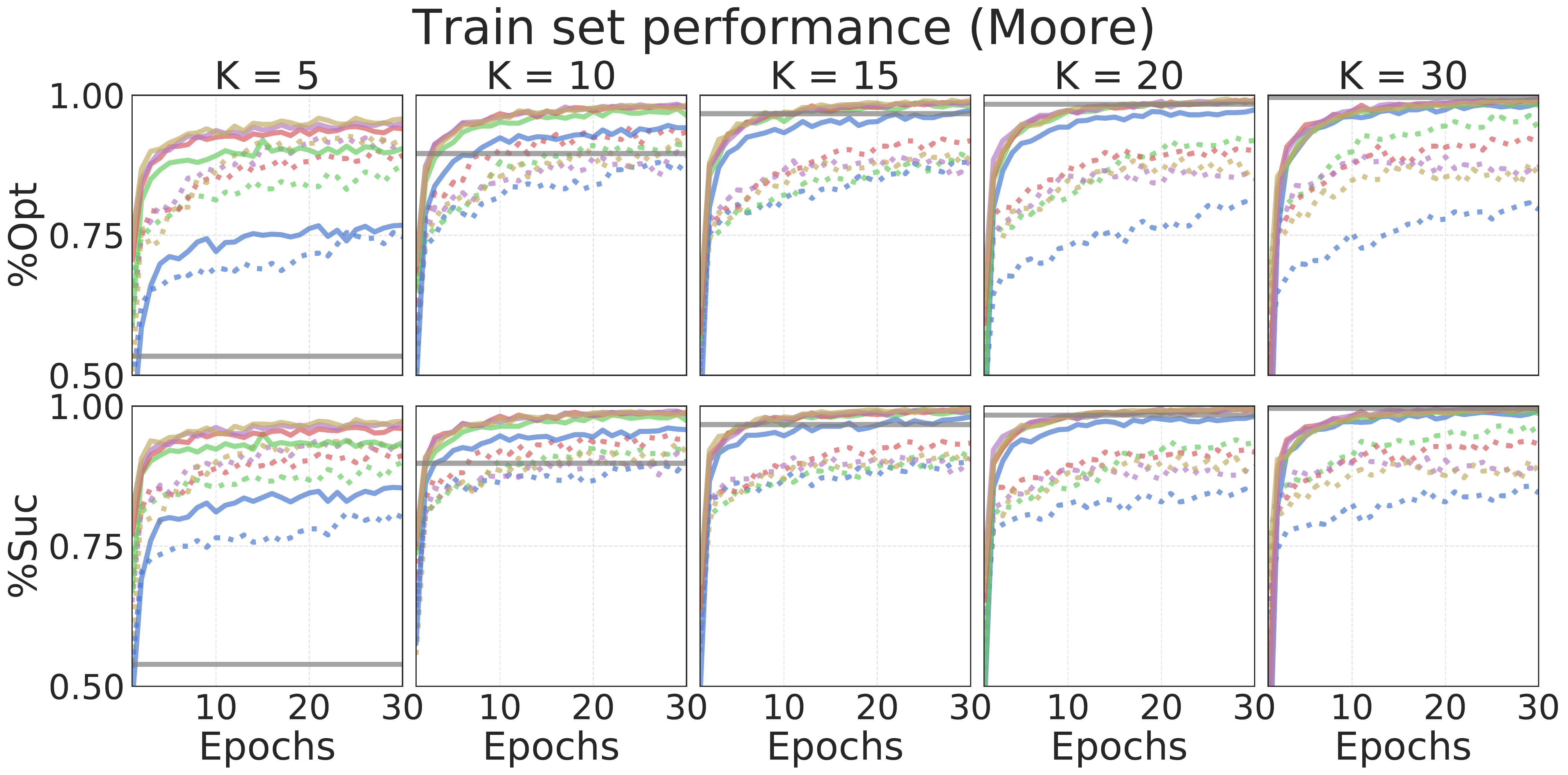}
\includegraphics[height=\figheight, trim={125pt 0 0 0}, clip]{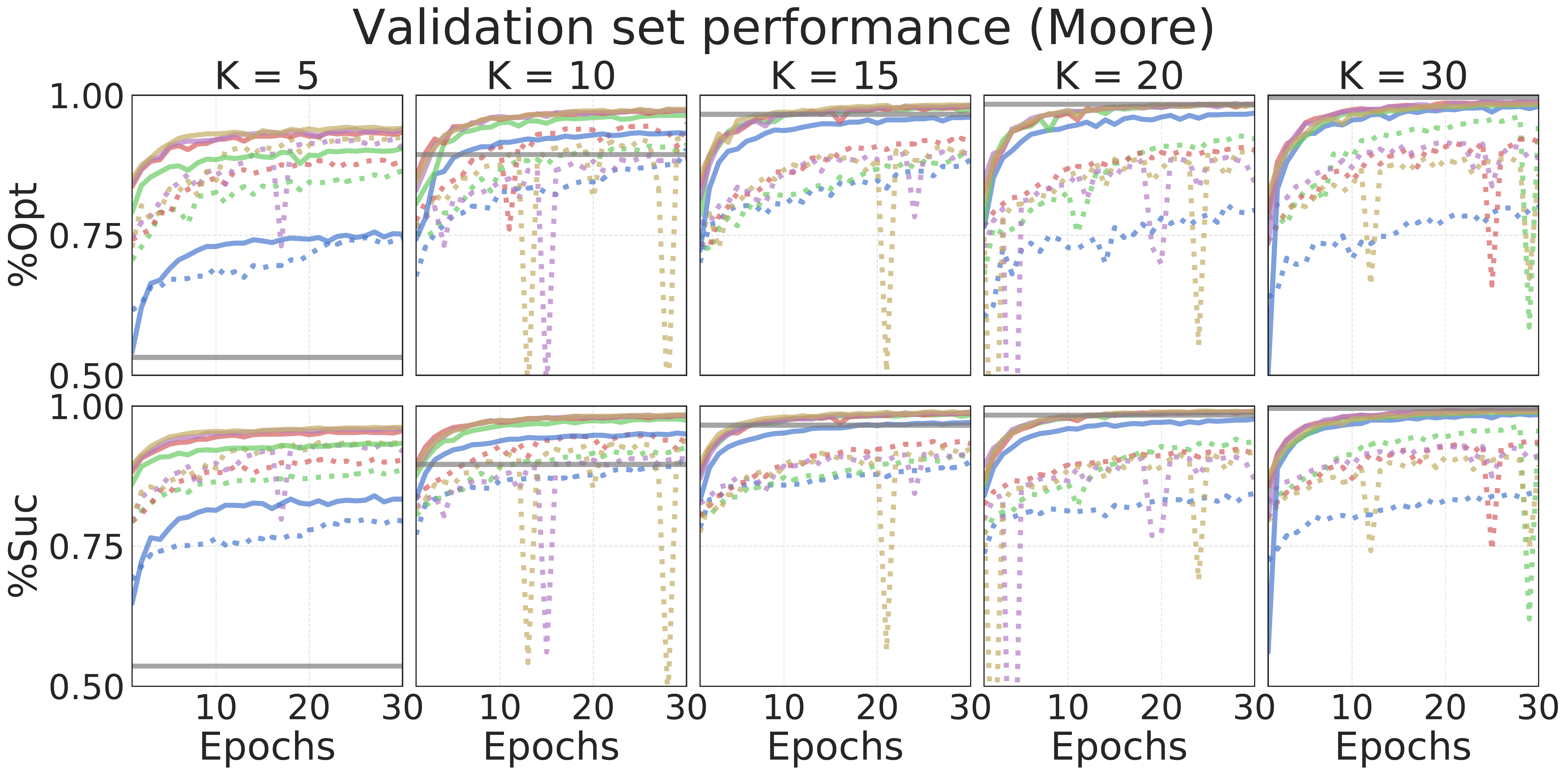}

\includegraphics[height=\figheight]{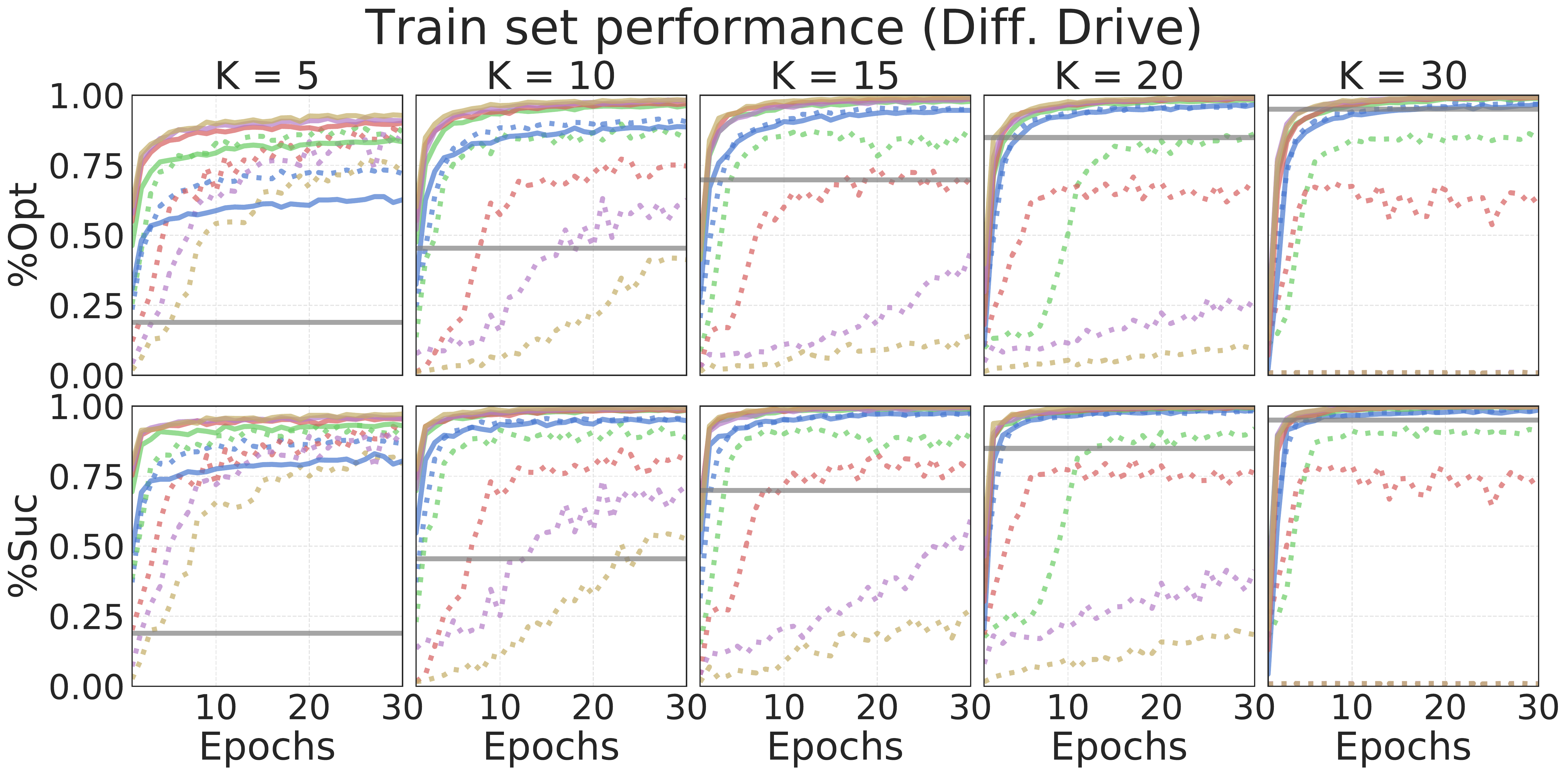}
\includegraphics[height=\figheight, trim={125pt 0 0 0}, clip]{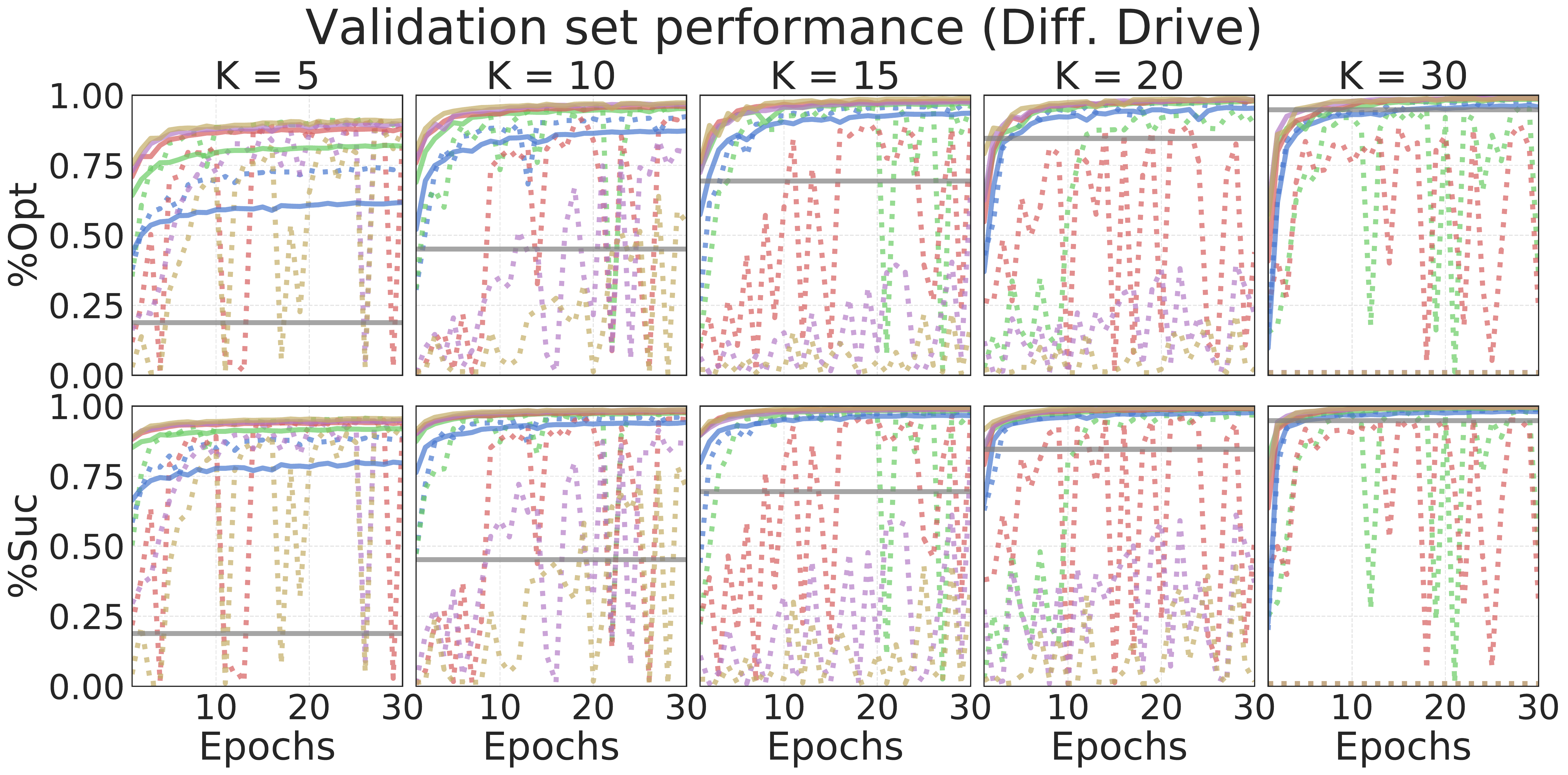}
\vspace{-0.1in}
\caption{Performance on 2D mazes of size $15 \times 15$ with varying iteration counts K and kernel sizes F . All models are trained using dataset size 25k. VIN exhibits higher training instability, its performance often oscillating between epochs. }
\label{fig:2d-vary-k}
\end{figure*}
\subsection{Random Seed and Hyperparameter Sensitivity}
\label{section:sensitivity}

The hypothesis this section sought to verify was whether the particular recurrent-convolutional form of the VIN did indeed negatively affect its optimization, as many ungated recurrent updates suffer from optimization problems including training instability and higher sensitivity to weight initialization and hyperparameters due to gradient scaling problems~\cite{PascanuMB13}.

We test each architecture's sensitivity to random seeds by running several experiments with the same hyperparameters but different random seeds, and measuring the variance in their final performance. These results are reported in Table~\ref{table:variance}. The results show that GPPN gets consistently lower variance than VIN over different random seed initializations, supporting the hypothesis that the LSTM update enables more training stability and easier optimization than the ungated recurrent update in VIN. 

We additionally test hyperparameter sensitivity in Figure~\ref{fig:hyperparameter-sensitivity}. We take all the results obtained on a hyperparameter sweep over settings $(K,F)$ where $K$ was varied over $K\in \{5,10,15,20,30\}$ and $F$ was varied over $F\in\{3,5,7,9,11\}$. We then rank these results, and the x-axis is the top-$x$ ranked hyperparameter settings and the corresponding y-axis is the average \%Opt/\%Suc of those $x$ settings. This plot thus measures how stable the performance of the architecture is to hyperparameter changes as the number of hyperparameter settings we consider grows. Therefore, architectures whose average top-$x$ ranked performance remains high and relatively flat demonstrates that good performance with the architecture can be obtained with many different hyperparameter settings. This suggests that these models are both easier to optimize and consistently better than alternatives, and higher performance was not due to a single lucky hyperparameter setting. We can see from the figures that the performance of GPPN is clearly both higher and more stable over hyperparameter settings than VIN.

In Figure~\ref{fig:2d-vary-k}, we plot the learning curves for VIN and GPPN on 2D mazes with varying $K$ and $F$. These plots show that VIN's performance often oscillates between epochs (especially for larger kernel sizes $F > 3$), while GPPN is much more stable. Learning curves for other experiments showing a similar result are included in the Appendix. The training stability of GPPN provides more evidence to the hypothesis that GPPNs are simpler to optimize than VINs and consistently outperform them.

%!TEX root = ../main.tex
 \begin{table}[t]
  \centering\footnotesize
\caption{The number of epochs it takes for each model to attain a certain \%Opt (50\%, 75\%, 90\%, 95\%) on the validation set under best settings of $(K, F)$. The results are on 2D mazes of size $15 \times 15$. GPPN learns faster.}
\vspace{2pt}
  \setlength\tabcolsep{3pt}
  \begin{tabular}{c|cccc|cccc|cccc}
    \toprule
     & \multicolumn{4}{c|}{NEWS} &  \multicolumn{4}{c|}{Moore} &  \multicolumn{4}{c}{Diff. Drive} \\
    Model & 50 & 75 & 90 & 95
      & 50 & 75 & 90 & 95
      & 50 & 75 & 90 & 95
    \\
    \midrule
VIN & 1 & 6 & 17 & -- & 1 & 1 & 11 & 23 & 2 & 3 & 5 & 14\\
GPPN & 1 & 1 & 3 & 5 & 1 & 1 & 3 & 5 & 1 & 2 & 3 & 6\\
    \bottomrule
    \end{tabular}
\label{table:2d-learning-speed}
\end{table}
\subsection{Learning Speed}

In this section, we examine whether VINs or GPPNs learn faster. To do this, we measure the number of training epochs (passes over the entire dataset) that it takes for each model to reach a specific \%Opt for the first time. These results are reported in Table~\ref{table:2d-learning-speed}. We can see from this table that GPPN learns significantly faster, often reaching 95\% within 5-6 epochs. Comparatively, VIN sometimes never reaches 95\%, as is the case for the NEWS mazes, or it takes 2-5 times as many epochs. This is the case even on the Differential Drive mazes, where VIN takes 2-3 times longer to train despite also getting high final performance. 

%!TEX root = ../main.tex
\begin{table}[t]
  \centering\footnotesize
\caption{Test performance on 2D mazes with \textbf{varying maze sizes $m \times m$} under best settings of $(K, F)$ for each model. For the larger $28 \times 28$ maze, we train for 100 epochs and sweep over $K \in \{14, 28, 56\}$ to account for longer trajectories required to solve some mazes. GPPN performs better.}
  \setlength\tabcolsep{3pt}
  \vspace{2pt}
  \begin{tabular}{cc|cc|cc|cc}
    \toprule
    & & \multicolumn{2}{c|}{NEWS} & \multicolumn{2}{c|}{Moore} & \multicolumn{2}{c}{Diff. Drive}\\
    $m$ & Model & \%Opt & \%Suc & \%Opt & \%Suc & \%Opt & \%Suc \\
    %Model & \%Opt & stdev & \%Suc & stdev & \%Opt & stdev & \%Suc & stdev\\ 
    \midrule
15 & VIN & 93.9 & 94.1 & 96.3 & 96.6 & 98.4 & 99.1\\
15 & GPPN & \textbf{99.2} & \textbf{99.5} & \textbf{98.8} & \textbf{99.3} & \textbf{99.3} & \textbf{99.7}\\
\midrule
28 & VIN & 93.0 & 93.2 & 95.0 & 95.8 & 93.8 & 96.8\\
28 & GPPN & \textbf{98.3} & \textbf{98.9} & \textbf{97.8} & \textbf{98.7} & \textbf{99.0} & \textbf{99.6}\\
    \bottomrule
    \end{tabular}
\label{table:2d-varying-maze-size}
\end{table}

\subsection{Larger Maze Size}

To test whether the improved performance GPPN persists even on larger, more challenging mazes, we evaluated the models on a dataset of mazes of size $28 \times 28$, and varied $K\in\{14,28,56\}$ (Table~\ref{table:2d-varying-maze-size}). We used a training dataset size of 25k. GPPN outperformed VIN by a significant margin (3-6\% for \%Opt and \%Suc) for all cases except Diff. Drive $15 \times 15$, where the gap was closer (GPPN 99.3 vs. VIN 98.3 for \%Opt).

%!TEX root = ../main.tex
\begin{table}[t]
  \centering\fontsize{8}{9}\selectfont
\caption{Performance on \textbf{3D ViZDoom mazes}. \%Acc is accuracy for predicting the top-down 2D maze design from first-person RGB maze images. When \%Acc is low, then the model must use a noisy maze design from which to plan, so \%Opt and \%Suc suffer as well. The results were obtained using $K=30$, the best setting of $F$ for each transition kernel, a smaller dataset size 10k (due to memory and time constraints), a smaller learning rate 5e-4, and 100 training epochs. VIN is more prone to overfitting: its validation \%Acc is low for all three transition kernels, while GPPN achieves higher validation \%Acc on NEWS and Moore.}
\vspace{2pt}
  \setlength\tabcolsep{1.5pt}
  \begin{tabular}{cc|ccc|ccc|cc}
    \toprule
    & & \multicolumn{3}{c|}{Train} & \multicolumn{3}{c|}{Val} & \multicolumn{2}{c}{Test}  \\
    Kernel & Model & \%Acc & \%Opt & \%Suc & \%Acc & \%Opt & \%Suc & \%Opt & \%Suc \\
    \midrule
NEWS & VIN & 99.9 & 82.3 & 83.0 & 81.5 & 80.8 & 81.5 & 79.0 & 79.7\\
NEWS & GPPN & 99.9 & 99.4 & 99.7 & 94.9 & 93.2 & 94.9 & 94.1 & 95.9\\
\midrule
Moore & VIN & 99.6 & 86.5 & 88.9 & 89.1 & 86.7 & 89.1 & 84.6 & 87.6\\
Moore & GPPN & 99.6 & 98.1 & 99.4 & 97.4 & 95.3 & 97.4 & 94.5 & 97.2\\
\midrule
Diff. Drive & VIN & 100.0 & 99.4 & 99.7 & 90.5 & 89.0 & 90.5 & 96.9 & 97.9\\
Diff. Drive & GPPN & 99.8 & 99.5 & 100.0 & 85.0 & 81.0 & 85.0 & 91.4 & 96.0\\
    \bottomrule
    \end{tabular}
\label{table:3d-results}
\vspace{-0.1in}
\end{table}

\subsection{3D ViZDoom Experiments}
\label{section:3d-ViZDoom}

In the 3D ViZDoom experiments, the state vector consists of RGB images showing the first-person view of the environment at each position and orientation, instead of the top-down 2D maze design (represented by a binary $m \times m$ matrix) as in the 2D maze experiments. To process the map images, we use a Convolutional Neural Network~\cite{lecun1989backpropagation} consisting of two convolutional layers: first layer with 32 filters of size $8 \times 8$ and a stride of 4, and second layer with 64 filters of size $4 \times 4$ with a stride $2 \times 2$, followed by a linear layer of size 256.\footnote{This architecture was adapted from a previous work which is shown to perform well at playing deathmatches in Doom~\cite{lample2017playing}.} The 256-dimensional representation for all the 4 orientations at each location is concatenated to create a 1024-dimensional representation. These representations of each location are then stacked at the corresponding x-y coordinate to create a map representation of size $1024\times m \times m$. The map representation is then passed through two more convolutional layers (first layer with 64 filters and the second layer with 1 filter, both of size $3 \times 3$ and a stride of 1) to predict a maze design matrix of size $1 \times m \times m$, which is trained using an auxillary binary cross-entropy loss. The predicted maze design is then stacked with the goal map and passed to the VIN or GPPN module in the same way as the 2D experiments.

The 3D ViZDoom results are summarized in Table~\ref{table:3d-results}. \textbf{\%Acc} is the accuracy for predicting the top-down 2D maze design from first-person RGB images. Learning to plan in the 3D environments is more challenging due to the difficulty of simultaneously optimizing both the original planner loss and the auxiliary maze prediction loss. We can see that when \%Acc is low, \ie, the planner module must rely on a noisy maze design, then the planner metrics \%Opt and \%Suc also suffer. We observe that VIN is more prone to overfitting on the training dataset: its validation \%Acc is low ($<91\%$) for all three transition kernels, whereas GPPN achieves higher validation \%Acc on NEWS and Moore. However, GPPN also overfits on the Differential Drive.
%!TEX root = ../main.tex
\section{Related Works}

\citet{qmdpnet} looked at extending differentiable planning towards being able to plan in partially observable environments. In their setting, the agent is not provided a-priori with its position within the environment and thus needs to maintain a belief state over where it actually is. Similar to VIN's differentiable extension of VI, the QMDP-Net architecture was based on creating a differentiable analogue of the QMDP algorithm~\citep{littman1995learning}, an algorithm designed to approximate belief space planning in POMDPs. The architecture itself consisted of a filter module, which maintained the beliefs over which states the agent currently was in, and a planning module, which determined what action to take next. The planning module was essentially using a VIN to enable it to make more informed decisions on which parts of the environment to explore.

In recent work there has been a variety of deep reinforcement learning models that have examined combining an internal planning process with model-free methods. The Predictron~\citep{silver2016predictron} was a value function approximator which predicted a policy's value by internally rolling out an LSTM forward predictive model of the agent's future rewards, discounts and values. These future rewards, values and discounts were then accumulated together, with the idea that this would predict a more accurate value by forcing the architecture to model a multi-step rollout. A later extension, Value Predictive Networks~\citep{oh2017value}, learnt a forward model that is used to predict the future rewards and values of executing a multi-step rollout. Although similar to the Predictron, they considered the control setting, where not only a value function had to be learnt but a policy as well. They demonstrated that their model, trained using model-free methods, was able to outperform existing methods on a 2D goal navigation task and outperformed DQN on Atari games. 

Convolutional-recurrent networks similar to the VIN and GPPN have had a recent history of use within computer vision, particularly for applications which have both a spatial and temporal aspect. Convolutional LSTMs (ConvLSTMs) were first used in the application of precipitation nowcasting, where the goal was to predict rainfall intensity within a region using past data~\cite{DBLP:conf/nips/ShiCWYWW15}. Recurrent-convolutional networks have also been used within computer vision applications where there is no explicit temporal aspect, such as object recognition. Feedback Networks \cite{feedbacknet} utilized a ConvLSTM in order to allow information to feedback from higher layers to lower layers by unrolling the ConvLSTM over time. This enabled the Feedback Network to attain performance better than or on par with Residual Networks (ResNets)~\cite{resnet}, one of the most commonly used feedforward architectures for object recognition.  

A deeper connection has also been explored between residual and convolutional-recurrent networks. \cite{bridgingthegap} tested whether weight tying between layers in a ResNet significantly affects performance, finding that although performance slightly degrades, the change is not drastic. They provide some hypotheses on these results, suggesting that deep feedforward networks like ResNets are approximating recurrent networks in some capacity. While the GPPN can be seen as an instance of ConvLSTMs, our paper is the first to apply it to the domain of differentiable path planning and to show that, in general, structuring differentiable path planning within the context of convolutional-recurrent networks enables use of previous well-established recurrent architectures such as LSTM and GRUs.

%!TEX root = ../main.tex

\section{Conclusion}

In this work, we re-formulated VIN as a convolutional-recurrent network and designed a new planning module called the Gated Path Planning Network (GPPN) which replaced the unconventional recurrent update in VIN with a well-established gated LSTM recurrent operator.
We presented experimental results comparing VIN and GPPN on 2D path-planning maze tasks and a 3D navigation task in the video game Doom, showing that the GPPN achieves results no worse and often better than VIN. The LSTM update alleviates many of the optimization issues including training instability and sensitivity to random seeds and hyperparameter settings. 
The GPPN is also able to utilize a larger kernel size, which the VIN is largely unable to do due to training instability, allowing the GPPN to work as well as VIN with fewer iterations. The GPPN also learns significantly faster, attaining high performance after only a few epochs, whereas the VIN takes longer to train. Finally, the relative performance improvement of GPPN over VIN increases with less training data. In conclusion, our analyses suggest that the inductive biases of VIN are not necessary in the design of a well-performing differentiable path planning module, and that the use of more general, gated recurrent architectures  provides significant benefits over VINs.

\section*{Acknowledgements}

LL is supported by a NSF GRFP Fellowship and by the CMU SEI under Contract FA8702-15-D-0002, Section H Clause, AFLCMC (H)-H001: 6-18014. EP, DC, and RS are supported in part by
Apple, Nvidia NVAIL, DARPA D17AP00001, IARPA DIVA award D17PC00340, and ONR award N000141512791. The authors would also like to thank Nvidia for providing GPU support.

\bibliography{main}

\begin{thebibliography}{22}
\providecommand{\natexlab}[1]{#1}
\providecommand{\url}[1]{\texttt{#1}}
\expandafter\ifx\csname urlstyle\endcsname\relax
  \providecommand{\doi}[1]{doi: #1}\else
  \providecommand{\doi}{doi: \begingroup \urlstyle{rm}\Url}\fi

\bibitem[Ackerman \& Guizzo(2015)Ackerman and Guizzo]{roomba}
Ackerman, E. and Guizzo, E.
\newblock irobot brings visual mapping and navigation to the roomba 980, 2015.

\bibitem[Chaplot et~al.(2018)Chaplot, Parisotto, and
  Salakhutdinov]{chaplot2018active}
Chaplot, D.~S., Parisotto, E., and Salakhutdinov, R.
\newblock Active neural localization.
\newblock In \emph{Proceedings of the 6th International Conference on Learning
  Representations (ICLR 2018)}, 2018.

\bibitem[Gupta et~al.(2017{\natexlab{a}})Gupta, Davidson, Levine, Sukthankar,
  and Malik]{cogmap}
Gupta, S., Davidson, J., Levine, S., Sukthankar, R., and Malik, J.
\newblock Cognitive mapping and planning for visual navigation.
\newblock In \emph{2017 {IEEE} Conference on Computer Vision and Pattern
  Recognition (CVPR 2017)}, pp.\  7272--7281, 2017{\natexlab{a}}.

\bibitem[Gupta et~al.(2017{\natexlab{b}})Gupta, Fouhey, Levine, and
  Malik]{unifyingmap}
Gupta, S., Fouhey, D.~F., Levine, S., and Malik, J.
\newblock Unifying map and landmark based representations for visual
  navigation.
\newblock \emph{CoRR}, abs/1712.08125, 2017{\natexlab{b}}.

\bibitem[Ha et~al.(2017)Ha, Dai, and Le]{DBLP:journals/corr/HaDL16}
Ha, D., Dai, A.~M., and Le, Q.~V.
\newblock Hypernetworks.
\newblock In \emph{Proceedings of the 5th International Conference on Learning
  Representations (ICLR 2017)}, 2017.

\bibitem[He et~al.(2016)He, Zhang, Ren, and Sun]{resnet}
He, K., Zhang, X., Ren, S., and Sun, J.
\newblock Deep residual learning for image recognition.
\newblock In \emph{2016 {IEEE} Conference on Computer Vision and Pattern
  Recognition (CVPR 2016)}, pp.\  770--778, 2016.

\bibitem[Hochreiter \& Schmidhuber(1997)Hochreiter and
  Schmidhuber]{hochreiter1997long}
Hochreiter, S. and Schmidhuber, J.
\newblock Long short-term memory.
\newblock \emph{Neural computation}, 9\penalty0 (8):\penalty0 1735--1780, 1997.

\bibitem[Karkus et~al.(2017)Karkus, Hsu, and Lee]{qmdpnet}
Karkus, P., Hsu, D., and Lee, W.~S.
\newblock {QMDP}-{N}et: Deep learning for planning under partial observability.
\newblock In \emph{Advances in Neural Information Processing Systems (NIPS
  2017)}, pp.\  4697--4707, 2017.

\bibitem[Kempka et~al.(2016)Kempka, Wydmuch, Runc, Toczek, and
  Ja\'skowski]{Kempka2016ViZDoom}
Kempka, M., Wydmuch, M., Runc, G., Toczek, J., and Ja\'skowski, W.
\newblock {ViZDoom}: A {D}oom-based {AI} research platform for visual
  reinforcement learning.
\newblock In \emph{IEEE Conference on Computational Intelligence and Games},
  pp.\  341--348. IEEE, Sep 2016.

\bibitem[Lample \& Chaplot(2017)Lample and Chaplot]{lample2017playing}
Lample, G. and Chaplot, D.~S.
\newblock Playing fps games with deep reinforcement learning.
\newblock In \emph{AAAI}, pp.\  2140--2146, 2017.

\bibitem[LeCun et~al.(1989)LeCun, Boser, Denker, Henderson, Howard, Hubbard,
  and Jackel]{lecun1989backpropagation}
LeCun, Y., Boser, B., Denker, J.~S., Henderson, D., Howard, R.~E., Hubbard, W.,
  and Jackel, L.~D.
\newblock Backpropagation applied to handwritten zip code recognition.
\newblock \emph{Neural computation}, 1\penalty0 (4):\penalty0 541--551, 1989.

\bibitem[Liao \& Poggio(2016)Liao and Poggio]{bridgingthegap}
Liao, Q. and Poggio, T.
\newblock Bridging the gaps between residual learning, recurrent neural
  networks and visual cortex.
\newblock \emph{arXiv preprint arXiv:1604.03640}, 2016.

\bibitem[Littman et~al.(1995)Littman, Cassandra, and
  Kaelbling]{littman1995learning}
Littman, M.~L., Cassandra, A.~R., and Kaelbling, L.~P.
\newblock Learning policies for partially observable environments: Scaling up.
\newblock In \emph{Machine Learning Proceedings 1995}, pp.\  362--370.
  Elsevier, 1995.

\bibitem[{Maze Generation Algorithms}(2018)]{wiki}
{Maze Generation Algorithms}.
\newblock Maze generation algorithms --- {W}ikipedia{,} the free encyclopedia,
  2018.
\newblock URL
  \url{https://en.wikipedia.org/wiki/Maze_generation_algorithm#Recursive_backtracker}.
\newblock [Online; accessed 9-Feb-2018].

\bibitem[Oh et~al.(2017)Oh, Singh, and Lee]{oh2017value}
Oh, J., Singh, S., and Lee, H.
\newblock Value prediction network.
\newblock In \emph{Advances in Neural Information Processing Systems}, pp.\
  6120--6130, 2017.

\bibitem[Pascanu et~al.(2013)Pascanu, Mikolov, and Bengio]{PascanuMB13}
Pascanu, R., Mikolov, T., and Bengio, Y.
\newblock On the difficulty of training recurrent neural networks.
\newblock In \emph{Proceedings of the 30th International Conference on Machine
  Learning (ICML 2013)}, pp.\  1310--1318, 2013.

\bibitem[Shi et~al.(2015)Shi, Chen, Wang, Yeung, kin Wong, and chun
  Woo]{DBLP:conf/nips/ShiCWYWW15}
Shi, X., Chen, Z., Wang, H., Yeung, D.-Y., kin Wong, W., and chun Woo, W.
\newblock Convolutional {LSTM} network: {A} machine learning approach for
  precipitation nowcasting.
\newblock In \emph{Advances in Neural Information Processing Systems (NIPS
  2015)}, pp.\  802--810, 2015.

\bibitem[Silver(2005)]{silver2005cooperative}
Silver, D.
\newblock Cooperative pathfinding.
\newblock In \emph{AIIDE}, pp.\  117--122, 2005.

\bibitem[Silver et~al.(2016)Silver, van Hasselt, Hessel, Schaul, Guez, Harley,
  Dulac-Arnold, Reichert, Rabinowitz, Barreto, et~al.]{silver2016predictron}
Silver, D., van Hasselt, H., Hessel, M., Schaul, T., Guez, A., Harley, T.,
  Dulac-Arnold, G., Reichert, D., Rabinowitz, N., Barreto, A., et~al.
\newblock The predictron: End-to-end learning and planning.
\newblock \emph{arXiv preprint arXiv:1612.08810}, 2016.

\bibitem[Sutton \& Barto(2018)Sutton and Barto]{Sutton:1998:IRL:551283}
Sutton, R. and Barto, A.
\newblock \emph{Introduction to Reinforcement Learning}.
\newblock MIT Press, 2nd edition, 2018.

\bibitem[Tamar et~al.(2017)Tamar, Wu, Thomas, Levine, and Abbeel]{vin}
Tamar, A., Wu, Y., Thomas, G., Levine, S., and Abbeel, P.
\newblock {V}alue {I}teration {N}etworks.
\newblock In \emph{Proceedings of the Twenty-Sixth International Joint
  Conference on Artificial Intelligence ({IJCAI} 2017)}, pp.\  4949--4953,
  2017.

\bibitem[Zamir et~al.(2017)Zamir, Wu, Sun, Shen, Shi, Malik, and
  Savarese]{feedbacknet}
Zamir, A.~R., Wu, T.-L., Sun, L., Shen, W.~B., Shi, B.~E., Malik, J., and
  Savarese, S.
\newblock Feedback networks.
\newblock In \emph{2017 IEEE Conference on Computer Vision and Pattern
  Recognition (CVPR)}, pp.\  1808--1817. IEEE, 2017.

\end{thebibliography}
\bibliographystyle{icml2018}

\clearpage

\begin{appendix}
%!TEX root = ../main.tex

\section{Learning Plots}

\begin{figure*}
\newcommand{\figheight}{250pt}
\centering
\includegraphics[width=0.83\textwidth, trim={0 0 270pt 0}, clip]{figures/pdf/legend_2rows.pdf}
\includegraphics[height=\figheight]{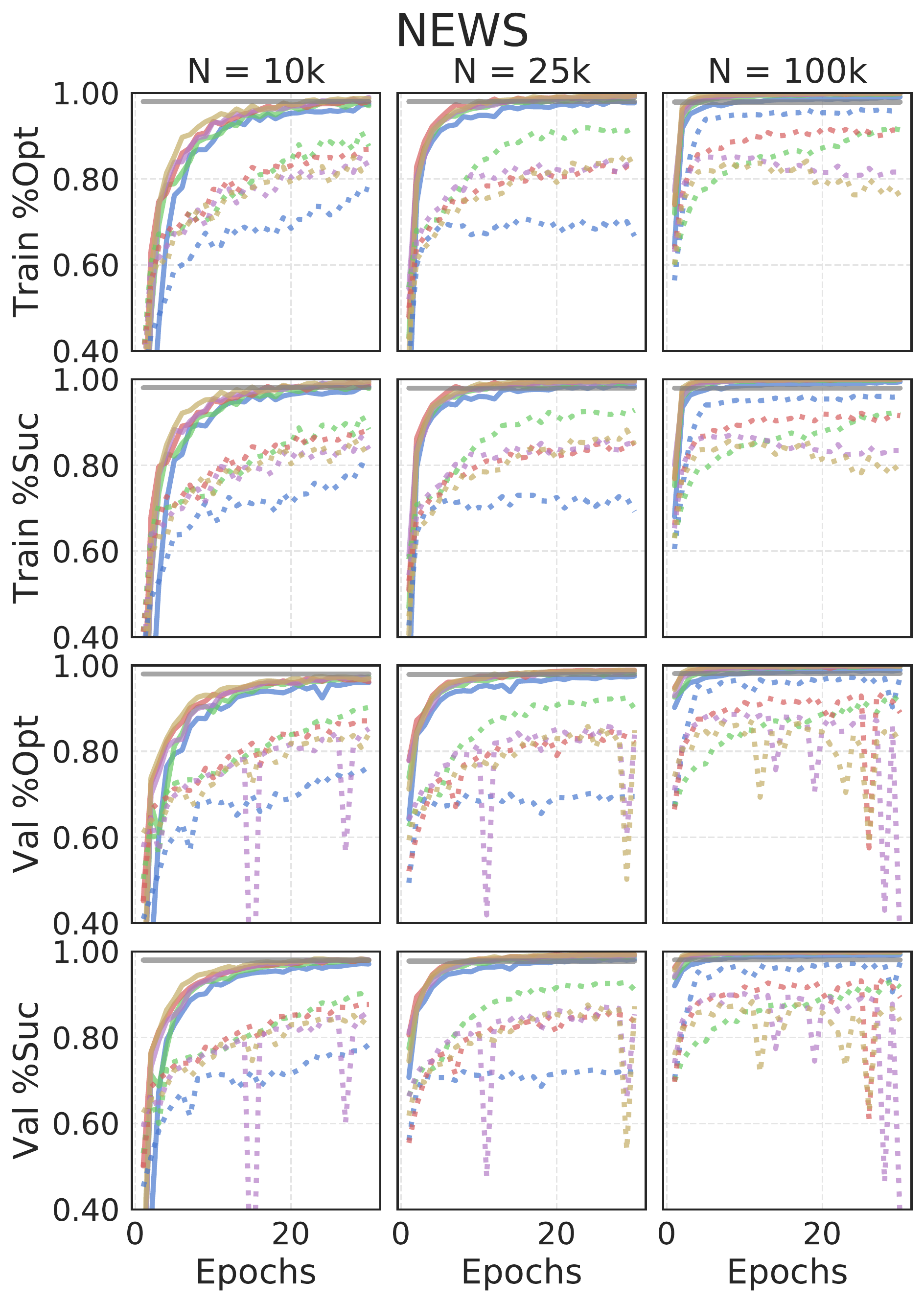}
\includegraphics[height=\figheight, trim={77pt 0 0 0}, clip]{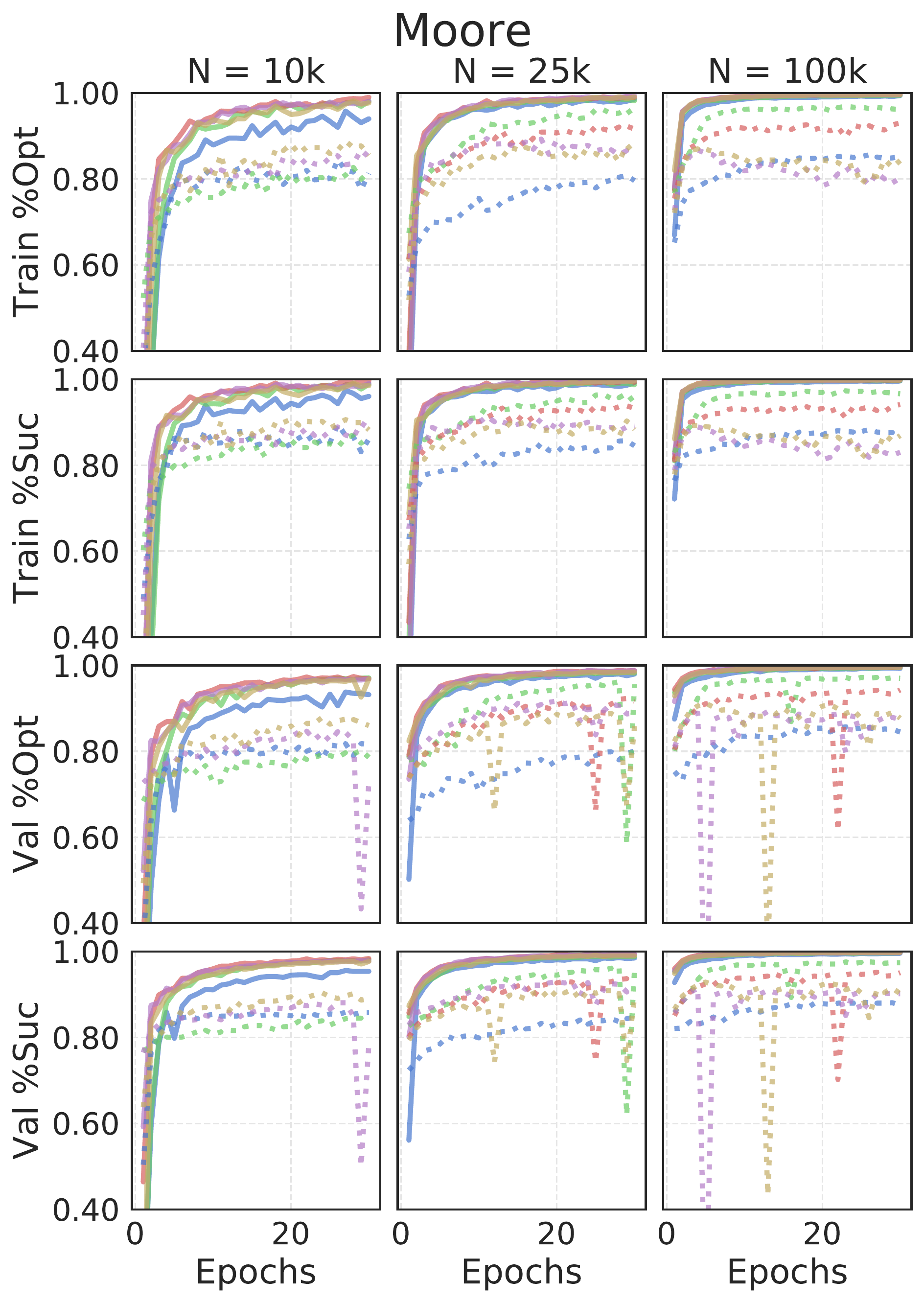}
\includegraphics[height=\figheight, trim={77pt 0 0 0}, clip]{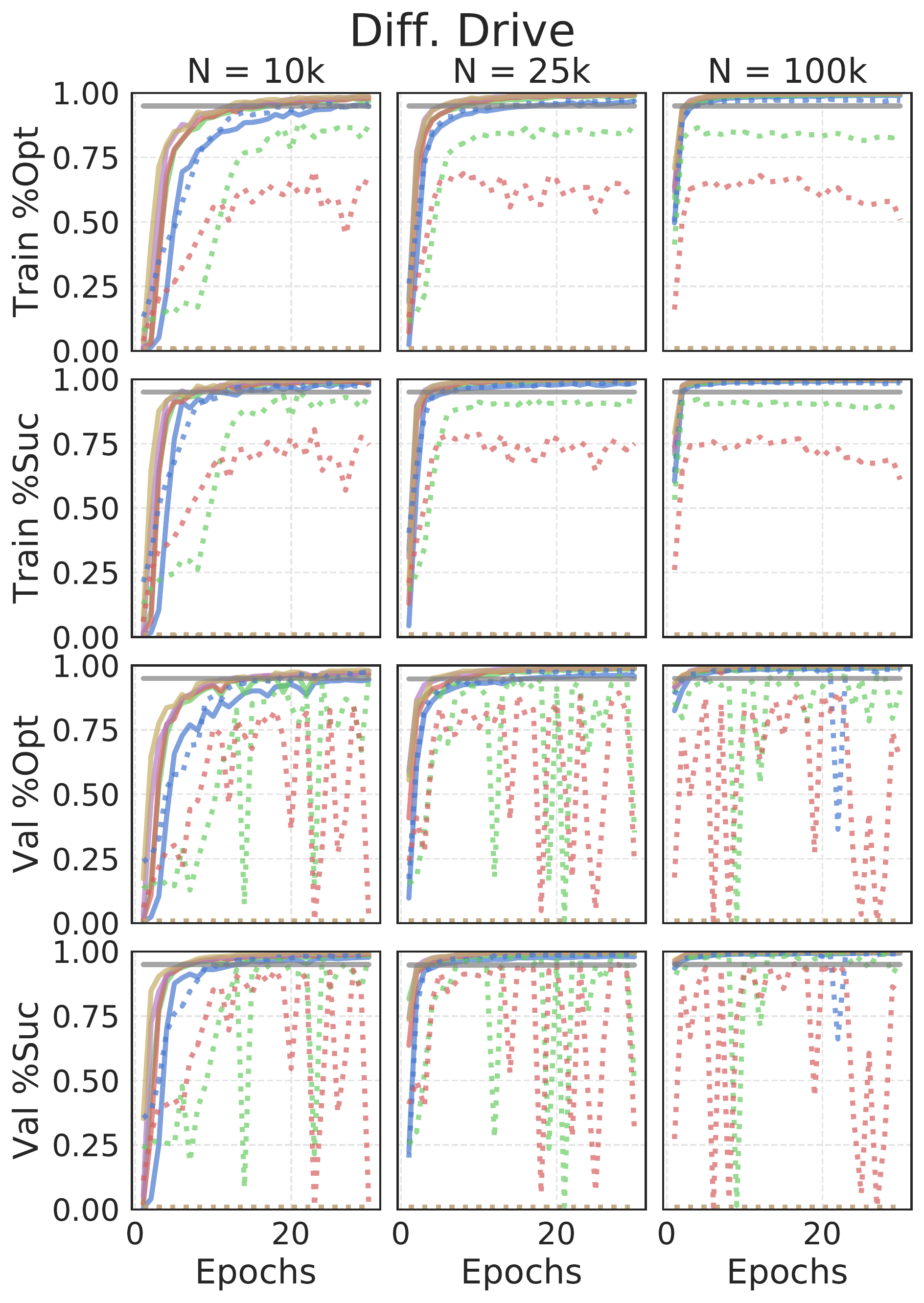}
\caption{ Performance on 2D mazes of size $15 \times 15$ with \textbf{varying dataset sizes} N. All models are trained using $K=30$ and learning rate 1e-3.}
\label{fig:train-size}
\end{figure*}

\begin{figure*}
\newcommand{\figheight}{250pt}
\centering
\includegraphics[height=\figheight]{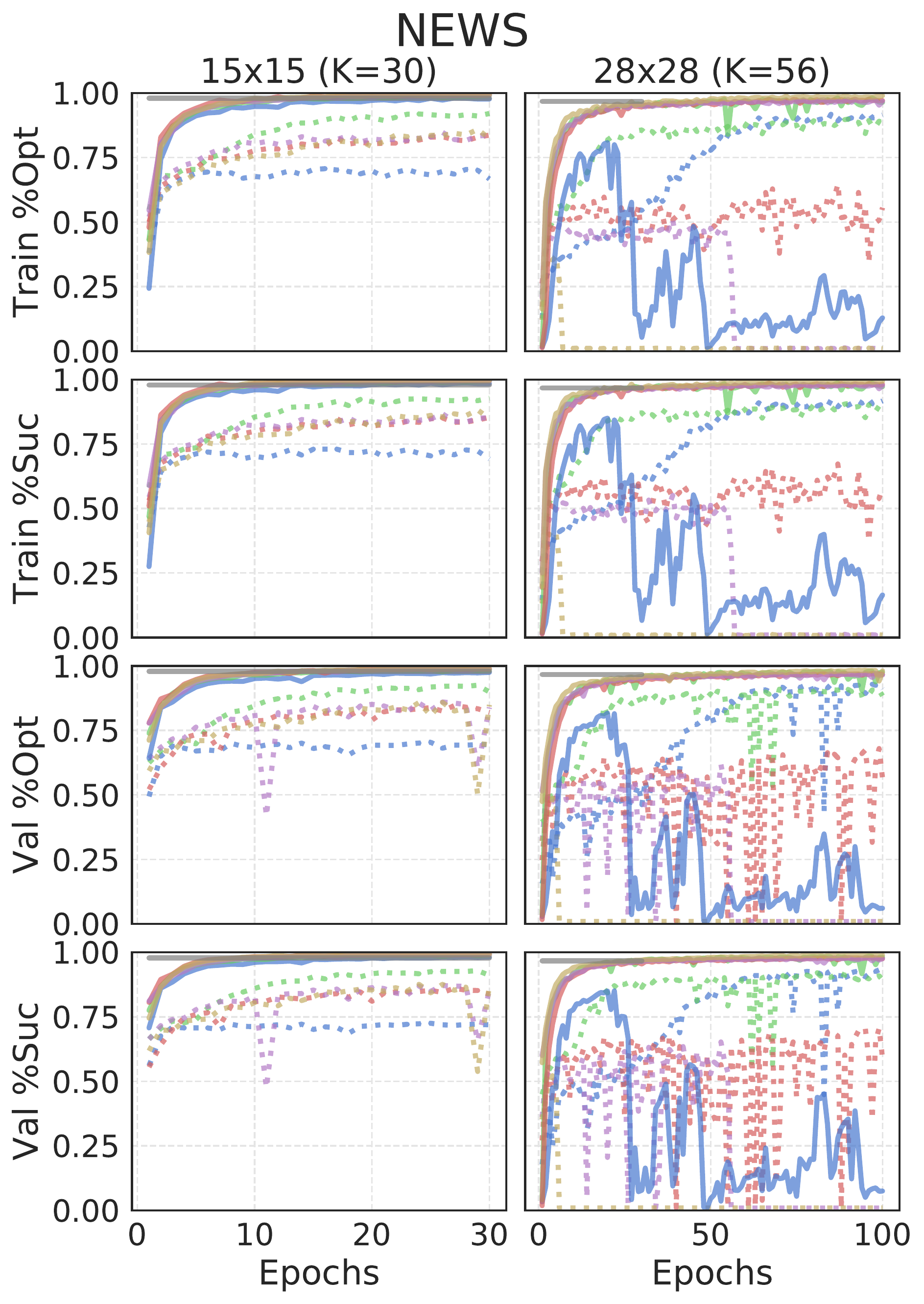}
\includegraphics[height=\figheight, trim={75pt 0 0 0}, clip]{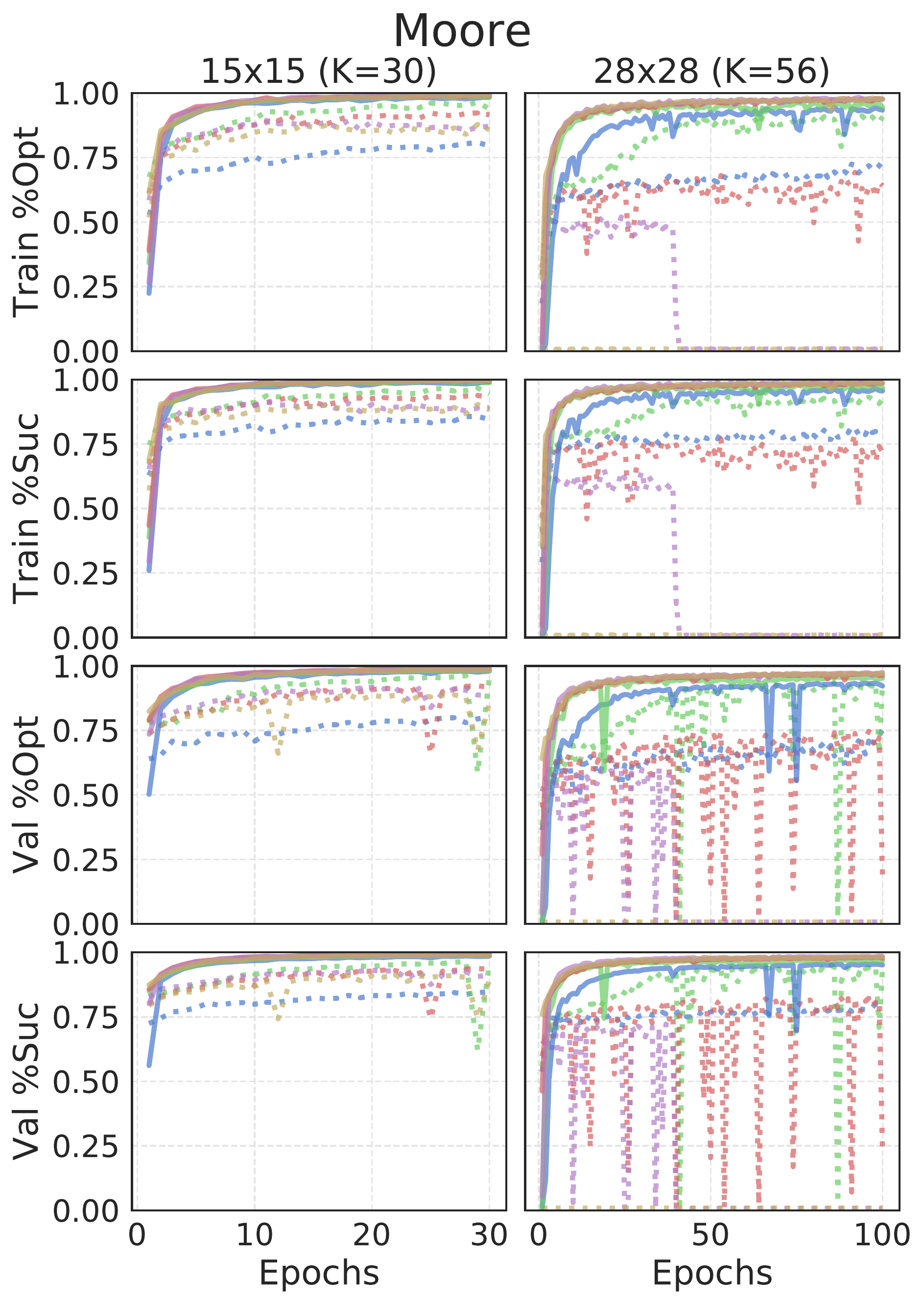}
\includegraphics[height=\figheight, trim={75pt 0 0 0}, clip]{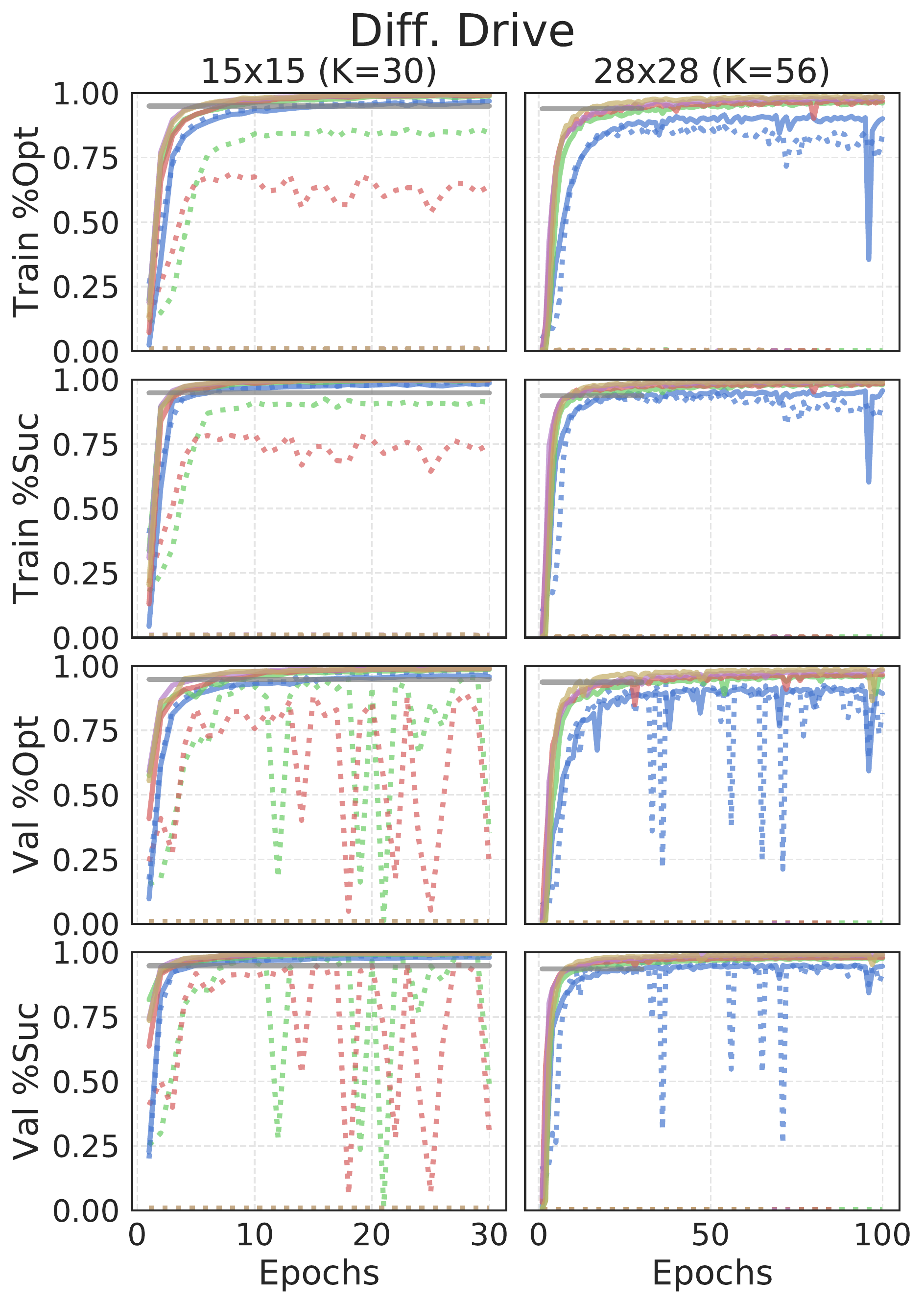}
\caption{Performance on 2D mazes with \textbf{varying maze sizes} $m \times m$. All models are trained using learning rate 1e-3, dataset size 25k, and $K=30$ (for $m=15$) or K=56 (for $m=28$).}
\label{fig:maze-size}
\end{figure*}

\begin{figure*}[t]
\newcommand{\figheight}{133pt}
\centering
\includegraphics[width=0.83\textwidth, trim={0 0 270pt 0}, clip]{figures/pdf/legend_2rows.pdf}
\includegraphics[height=\figheight]{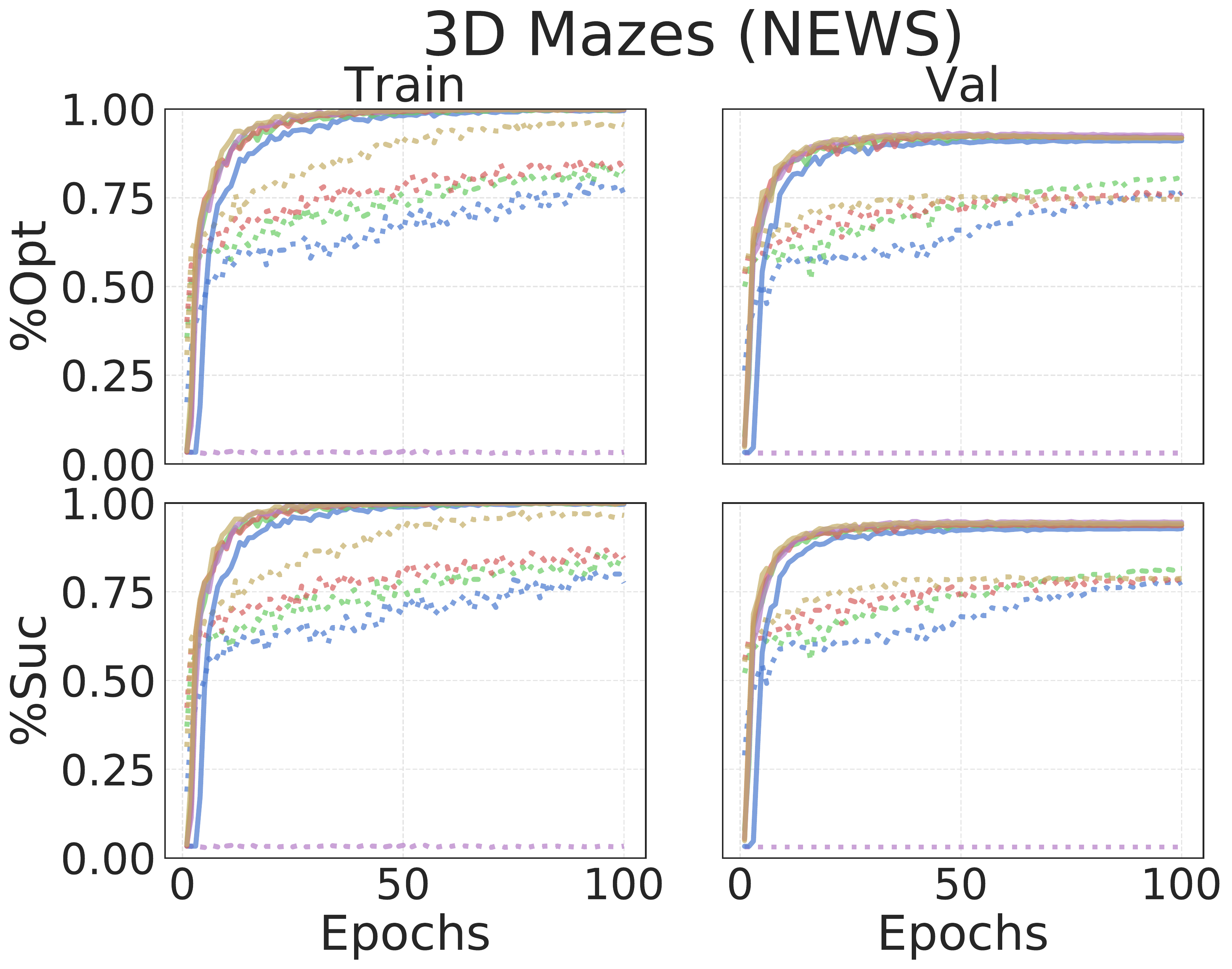}
\hspace{5pt}
\includegraphics[height=\figheight, trim={123pt 0 0 0}, clip]{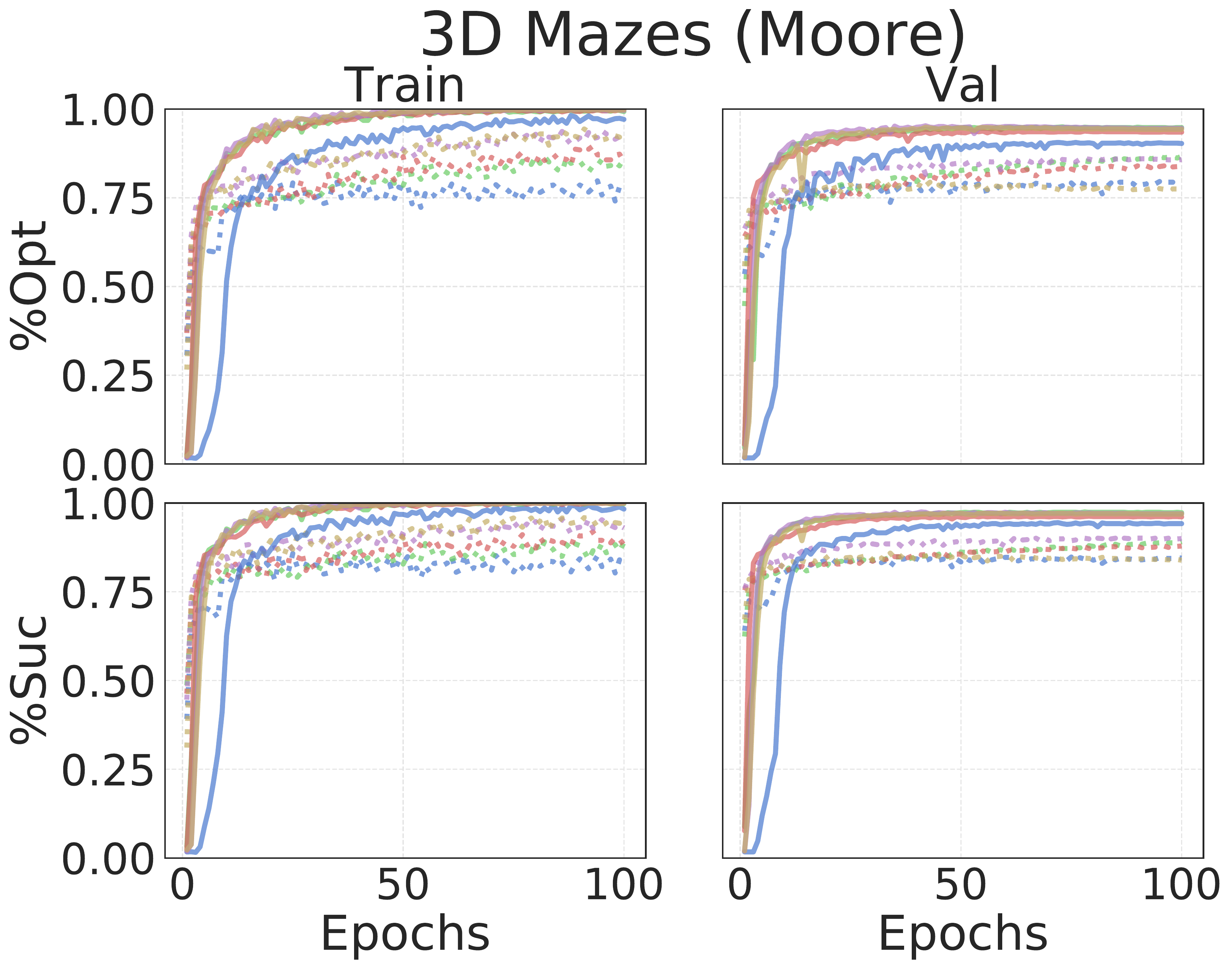}
\hspace{5pt}
\includegraphics[height=\figheight, trim={123pt 0 0 0}, clip]{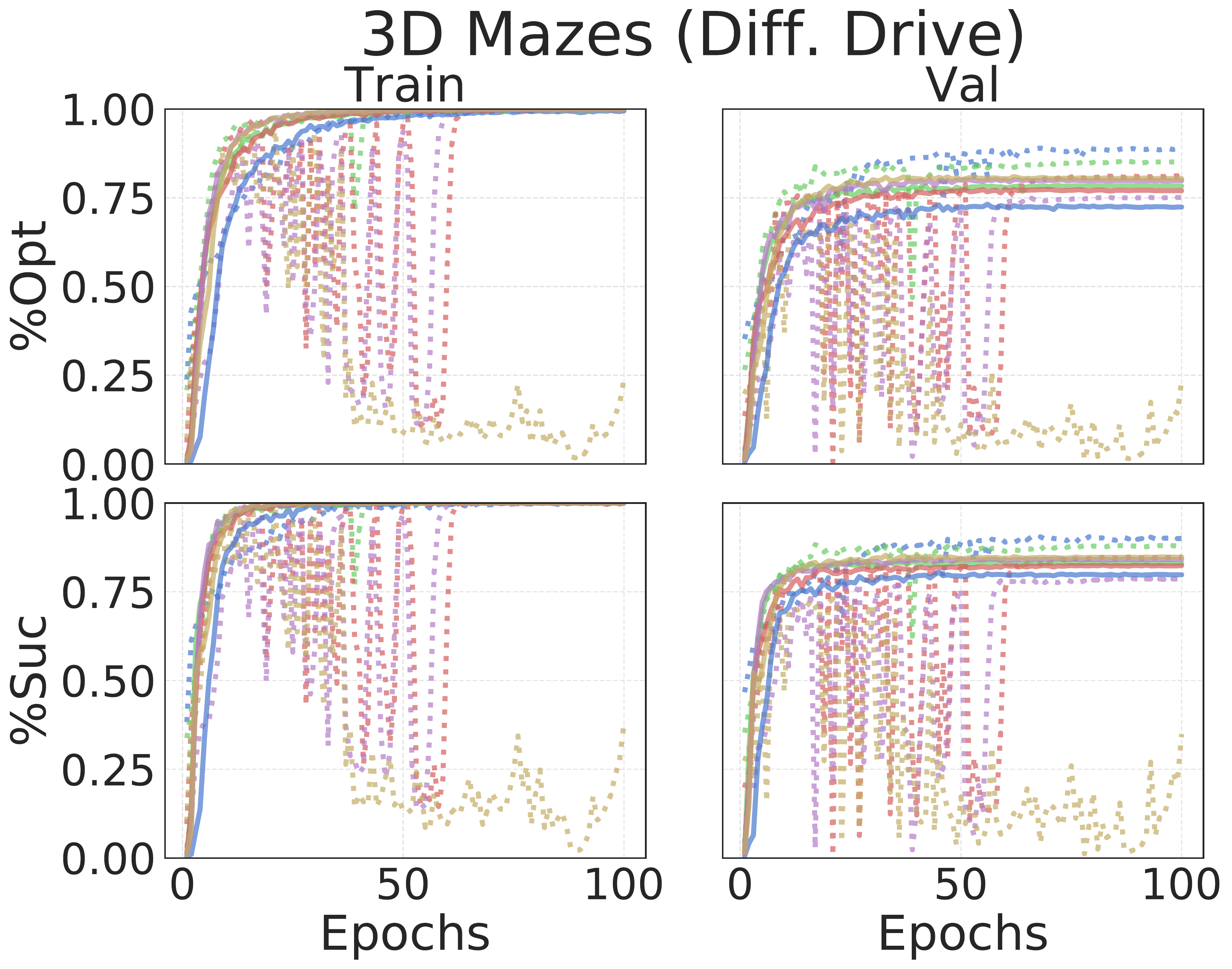}
\caption{Performance on 3D ViZDoom mazes of size $15 \times 15$. All models are trained using K=30, learning rate 5e-4, and dataset size 10k.}
\label{fig:3d-results}
\end{figure*}

As mentioned in Section~\ref{section:sensitivity}, we provide additional learning plots for varying dataset sizes (Figure~\ref{fig:train-size}), varying maze sizes (Figure~\ref{fig:maze-size}), and 3D ViZDoom results (Figure~\ref{fig:3d-results}).

\section{Hyperparameter Settings}
\label{section:hyperparameter-settings}

Unless otherwise noted, all models in Tables \ref{table:2d-varying-f}, \ref{table:2d-varying-k-f}, \ref{table:2d-varying-k}, \ref{table:2d-varying-train-size}, \ref{table:variance}, \ref{table:2d-learning-speed}, \ref{table:2d-varying-maze-size},
\ref{table:3d-results}
are trained on 2D mazes of size $15 \times 15$ for 30 epochs using a learning rate of 1e-3, batch size 32, gradient clipping of 40, and 25k/5k/5k train-val-test split. An initial sweep of learning rates over \{ 1e-4, 1e-3, 5e-3, 1e-2, 1e-1 \} found that GPPN is more robust to varying learning rates and that 1e-3 worked best for both models (see Table~\ref{table:lr}). We do a hyperparameter sweep of $(K, F)$ over $K \in \{5, 10, 15, 20, 30\}$ and $F \in \{3, 5, 7, 9, 11\}$. The only exceptions are the following: For the larger $28 \times 28$ maze in Table~\ref{table:2d-varying-maze-size}, we sweep $K$ over $\{14, 28, 56\}$ to account for longer trajectories required to solve some mazes. For the 10k and 100k dataset sizes in Table~\ref{table:2d-varying-train-size}, we used a train-test-val split of 10k/2k/2k and 100k/10k/10k, respectively. The variance results in Table~\ref{table:variance} are obtained using dataset size 100k. The 3D ViZDoom results in Table~\ref{table:3d-results} were obtained using $K=30$, the best setting of $F$ for each transition kernel, a smaller dataset size 10k , a smaller learning rate 5e-4, and 100 training epochs.

The particular form of the LSTM update the GPPNs take in the experimental section is slightly different from the standard one. The first difference is that we remove the dependence of each layer of the ConvLSTM on the "reward" function $\bar{R}_{[i',j',F]}$ since we did not find this skip-connection helped performance much in preliminary experiments. Therefore layers of the GPPN only take as input the previous layer's hidden units. The second change was made to make the GPPN easier to implement in a framework where built-in LSTM updates are available but ConvLSTMs are not. It first takes the convolution over the previous hidden layers and produces a 1-channel feature map, and then for each position passes that feature map position to the framework's built-in LSTM update. This is similar to having a single shared input gate for all the inputs. When tested against a GPPN with the standard LSTM update equation with full input gating, we did not observe any significant different in test metrics but the single input gate did save some computation time.

%!TEX root = ../main.tex

\begin{figure}[t]
\centering
\hspace{-0.2in}\includegraphics[width=0.5\linewidth]{figures/pdf/vin_maze.pdf}
\includegraphics[width=0.5\linewidth]{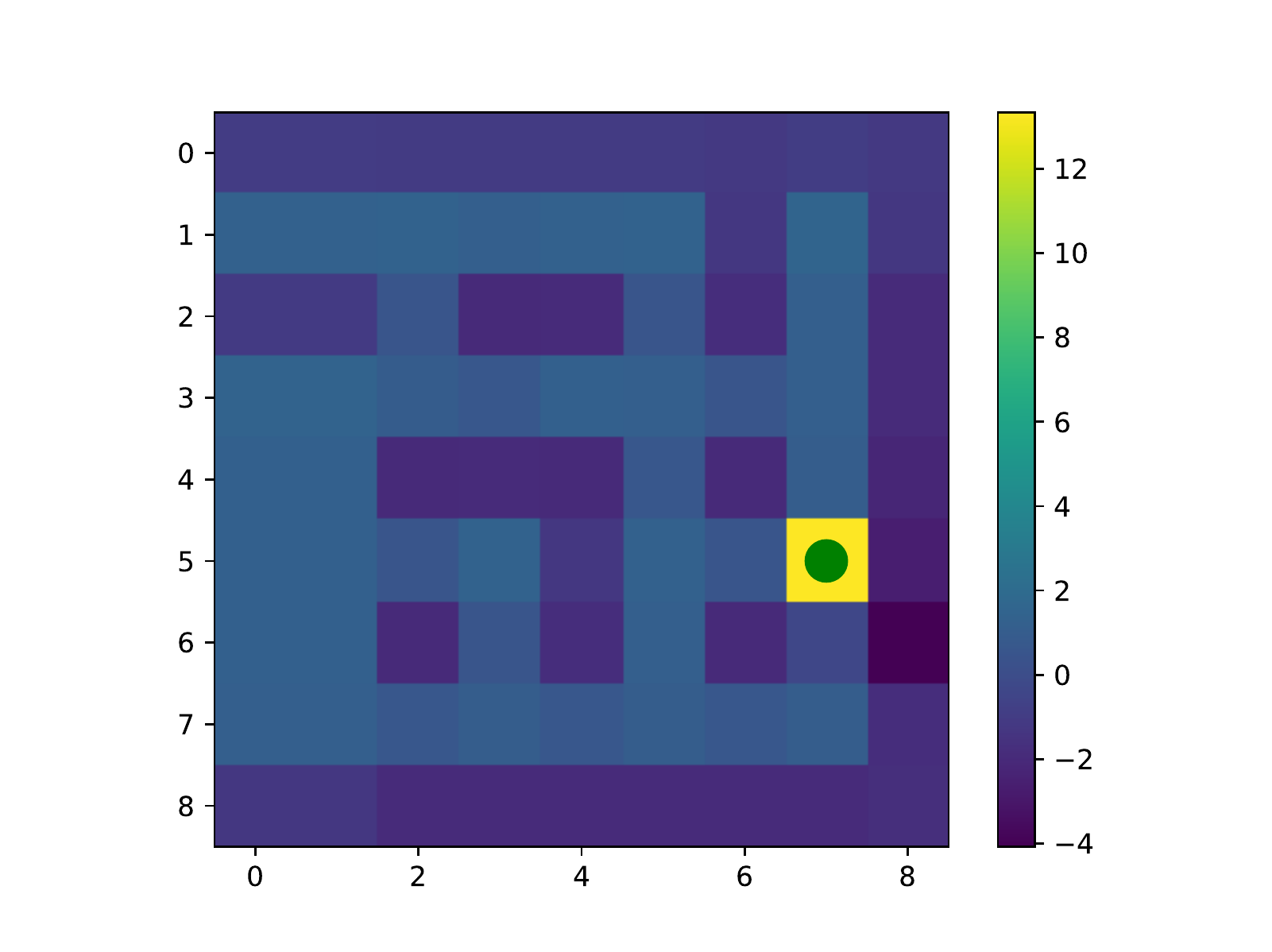}
\caption{\textbf{Left}: A sample 2D maze environment, where yellow cells, purple cells, and the green circle represent open spaces, walls, and the goal state respectively. All mazes are constructed as fully connected trees with a decimation parameter that destroys walls with a certain probability.  \textbf{Right}: The initial reward vector learned for the 2D maze task on a fully trained VIN. VIN gets around the spatial invariance of its model by applying a large negative reward to states that should never be entered (walls) and a large reward for the goal location.}
\label{fig:vinreward}
\end{figure}

\section{Hyper-VIN}

The VIN uses convolutions to represent the model, which causes it to effectively be spatially invariant, meaning VINs are incapable of truly solving mazeworld in the same way as value iteration on the true model. The result is that VINs learn a workaround that enables it to deal with non-linearities over the state space: it assigns a large negative reward to every wall position. This is shown in Figure~\ref{fig:vinreward}: the large reward gradient between walls and non-walls discourages the model from producing policies that ``visit'' wall states which would be impossible under the true model. Additionally, the spatial convolution model is fixed and invariant for all mazes, which is suboptimal as each MDP in the 2D environments require a different transition kernel based on the maze design.

In this section, we try to alleviate this issue by, first, untying the weights of the spatial convolution and, second, predicting the untied convolution weights directly from the maze design. We call this variant the Hyper-VIN, adopting the naming convention from HyperNetworks~\cite{DBLP:journals/corr/HaDL16} which also used the kernel of using a network with weights predicted from another network. To implement the Hyper-VIN, we predict for each position $(i,j)$ in the environment a convolutional weight matrix from the input map design. The Hyper-VIN update equation then becomes:
\begin{align*}
  \bar{V}^{(t)}_{i',j'} &= \omega\left(W_R^{\bar{a},i',j'} \bar{R}_{[i',j',3]} + W_V^{\bar{a},i',j'} \bar{V}^{(t-1)}_{[i',j',3]}\right)
\end{align*}

A question that can be asked about Hyper-VIN is if they perform as well as (or better than) the actual algorithms they were designed to mimic because the true algorithm is within the model class. This would provide some evidence whether such modules were actually computing the value, or whether they simply acted like recurrent networks and computed a less interpretable internal representation. Empirically, we instead found that Hyper-VINs have high variance in training and are difficult to optimize (see Table~\ref{table:hypervin}). Hyper-VINs trained by SGD often fail to reach the performance of their exact algorithmic counterpart (value iteration) on small mazes even though value iteration is within the hypothesis class of these models, suggesting that the optimization of such architectures is significantly difficult.

%!TEX root = ../main.tex

%!TEX root = ../main.tex

%!TEX root = ../main.tex

%!TEX root = ../main.tex
\begin{table*}[t]
  \centering
\caption{Test performance (\%Opt) on 2D mazes of size $15 \times 15$ with varying learning rates. The models were trained using dataset size 25K and the best $(K, F)$ settings for each maze transition kernel. ``--'' indicates the training diverged. Learning rate 1e-3 worked best for all models and transition kernels. GPPN is less sensitive to learning rate changes.}
  \setlength\tabcolsep{3pt}
  \vspace{2pt}
  \begin{tabular}{cccc|cccc}
    \toprule
    &&&& \multicolumn{4}{c}{\%Opt with learning rate} \\
Kernel & Model & $K$ & $F$ & 1e-4 & 1e-3 & 5e-3 & 1e-2 \\
\midrule
NEWS & VIN & 20 & 5 & 64.1 & \textbf{92.0} & 38.1 & 2.9\\
NEWS & GPPN & 20 & 11 & 95.5 & \textbf{99.0} & 96.9 & 19.4\\
\midrule
Moore & VIN & 30 & 5 & 77.1 & \textbf{85.9} & 75.2 & 2.7\\
Moore & GPPN & 30 & 9 & 94.0 & \textbf{98.8} & 82.2 & 2.0\\
\midrule
Diff. Drive & VIN & 30 & 3 & 74.2 & \textbf{97.5} & -- & --\\
Diff. Drive & GPPN & 30 & 9 & 91.7 & \textbf{99.3} & 96.3 & 18.6\\
    \bottomrule
    \end{tabular}
\label{table:lr}
\end{table*}

%!TEX root = ../main.tex
\begin{table*}[t]
	\centering
\caption{Test performance (mean and standard deviation) on 2D mazes of size $15 \times 15$, taken over 7 runs on the same dataset. These results were attained using iteration count $K=20$ for all models, filter size $F=3$ for VIN and Hyper-VIN, and $F=11$ for GPPN. Due to GPU memory limitations with Hyper-VIN, all models were trained using half the hidden dimension compared to experiments in the main paper. Hyper-VIN has high variance in training and is difficult to optimize.}
\label{table:hypervin}
	\begin{tabular}{*9c}
    \toprule
    & \multicolumn{4}{c}{NEWS} & \multicolumn{4}{c}{Differential Drive}\\
    Model & \multicolumn{2}{c}{\%Opt} & \multicolumn{2}{c}{\%Suc} & \multicolumn{2}{c}{\%Opt} & \multicolumn{2}{c}{\%Suc} \\
    & mean & stdev & mean & stdev & mean & stdev & mean & stdev\\
    \midrule
    Hyper-VIN & 75.1 & 11.4
    	& 80.2 & 9.5
        & 77.6 & 1.9
        & 94.8 & 0.6 \\
    VIN & 85.8 & 6.6 & 87.1 & 5.7
    	& 97.8 & 0.1 & 98.7 & 0.1\\
    GPPN & 98.9 & 0.2 & 99.3 & 0.2 & 98.1 & 0.9 & 98.8 & 1.1 \\
    Value Iteration & 94.2 & -- & 94.2 & --
       & 85.1 & -- & 85.1 & -- \\
       \bottomrule
    \end{tabular}
\end{table*}
\end{appendix}

\end{document}